\documentclass[letterpaper]{article} 
\usepackage{aaai25}  
\usepackage{times}  
\usepackage{helvet}  
\usepackage{courier}  
\usepackage[hyphens]{url}  
\usepackage{graphicx} 
\usepackage{natbib}  
\usepackage{caption} 
\usepackage{amsmath}
\usepackage{amssymb}
\usepackage{booktabs} 
\usepackage{color}
\usepackage{multirow}
\usepackage{bbding}
\usepackage{pifont}
\usepackage{tikz}
\usepackage{etoolbox}
\newcommand{\circled}[2][]{\tikz[baseline=(char.base)]
    {\node[shape = circle, draw, inner sep = 1pt]
    (char) {\phantom{\ifblank{#1}{#2}{#1}}};%
    \node at (char.center) {\makebox[0pt][c]{#2}};}}
\robustify{\circled}
\usepackage{algorithm}
\usepackage{algorithmic}

\usepackage{newfloat}
\usepackage{listings}
\DeclareCaptionStyle{ruled}{labelfont=normalfont,labelsep=colon,strut=off} 
\floatstyle{ruled}
\newfloat{listing}{tb}{lst}{}
\floatname{listing}{Listing}
\title{HS-FPN: High Frequency and Spatial Perception FPN for Tiny Object Detection}
\author{
    Zican Shi\textsuperscript{\rm 1}, Jing Hu\textsuperscript{\rm 1\rm 2}\thanks{Corresponding Author}, Jie Ren\textsuperscript{\rm 1}, Hengkang Ye\textsuperscript{\rm 1}, Xuyang Yuan\textsuperscript{\rm 1}, 
    Yan Ouyang\textsuperscript{\rm 2}, Jia He\textsuperscript{\rm 2}, Bo Ji\textsuperscript{\rm 2}, Junyu Guo\textsuperscript{\rm 2}
    \\
}
\affiliations{
    \textsuperscript{\rm 1}Huazhong University of Science and Technology, Wuhan, China\\
    \textsuperscript{\rm 2}National Key Laboratory of Science and Technology on Multi-Spectral Information Processing, Wuhan, China \\
    \{misaka10043, by6040130, jieren, yehengkang, lucasxyyuan\}@hust.edu.cn\\
    \{oyy\_01, hugh120007, jibo\_2006, henry\_rocky\_1999\}@163.com
}

\usepackage{bibentry}

\begin{document}

\maketitle

\begin{abstract}
The introduction of Feature Pyramid Network (FPN) has significantly improved object detection performance. However, substantial challenges remain in detecting tiny objects, as their features occupy only a very small proportion of the feature maps. Although FPN integrates multi-scale features, it does not directly enhance or enrich the features of tiny objects. Furthermore, FPN lacks spatial perception ability. To address these issues, we propose a novel High Frequency and Spatial Perception Feature Pyramid Network (HS-FPN) with two innovative modules. First, we designed a high frequency perception module (HFP) that generates high frequency responses through high pass filters. These high frequency responses are used as mask weights from both spatial and channel perspectives to enrich and highlight the features of tiny objects in the original feature maps. Second, we developed a spatial dependency perception module (SDP) to capture the spatial dependencies that FPN lacks. Our experiments demonstrate that detectors based on HS-FPN exhibit competitive advantages over state-of-the-art models on the AI-TOD dataset for tiny object detection. 


\end{abstract}



\section{Introduction}\label{introduction}
Tiny Object Detection (TOD), a subtask of general object detection, focuses on detecting tiny-sized objects that AI-TOD \cite{aitod} defines as being less than $16\times16$ pixels. TOD plays an important role in scenarios such as traffic sign detection, scene monitoring, unmanned aerial vehicle analysis, pedestrian detection, autonomous driving, and maritime rescue.

Despite the development of deep learning bringing forth many outstanding object detection frameworks that have achieved impressive performance on regular-sized objects, the performance of these frameworks drops significantly when it comes to detecting tiny objects. Based on recent reviews and surveys on tiny object detection \cite{soda, tiny-review1, tiny-review2}, we infer that the performance decline can be attributed to three factors: tiny objects have limited usable features, their features are easily interfered with, and general network architectures do not adequately focus on tiny objects. Moreover, in recent years, large vision models \cite{CLIP2CL, regionClip} have mostly been studied only on the general COCO dataset \cite{coco}, without giving extra attention to tiny objects. Due to these reasons, tiny object detection has gradually become a challenging research direction in the field of computer vision.

Feature Pyramid Network (FPN) \cite{fpn} has already become an indispensable part of object detection models \cite{faster-rcnn}. FPN leverages multi-scale feature maps produced by the backbone network (typically ResNet \cite{resnet}), propagating the strong semantics of deep feature maps to shallow feature maps, significantly enhancing the information content of the shallow features.

Although experiments have shown that FPN significantly improves object detection performance, there are still some challenges when it comes to detecting tiny objects, as described below:

\begin{figure*}
\centering
\includegraphics[width=0.90\textwidth]{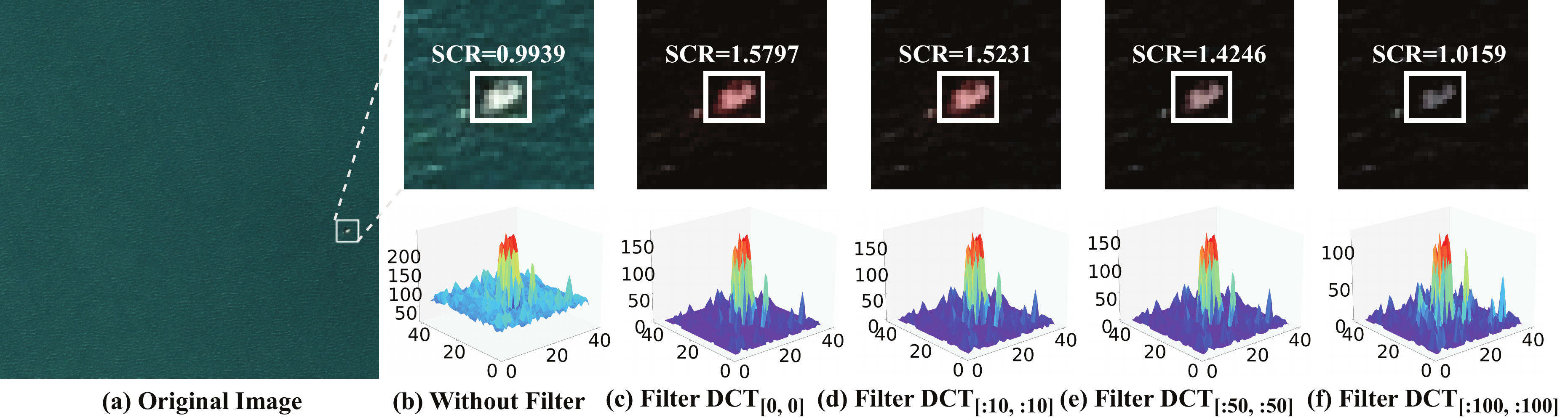} 
\caption{Target region after filtering out low-frequency components across different ranges (top), 3D surface representations (down), and the SCR of the target (marked in red text). (a) The original image containing a tiny ship target; (b) The target neighborhood image without filtering; (c - d) Filtering out low-frequency components in the top-left corner of the DCT results, from the $0\times0$ to $100\times100$ region. It shows that the SCR first increases and then decreases as the filtering range is expanded.}
\label{fig1}
\end{figure*}

\textbf{Tiny objects have limited usable features.} Although FPN integrates deep and shallow features, it does not directly increase the feature content for tiny objects. Due to the small pixel area of tiny objects, frequent downsampling by the backbone network continually compresses the feature size of tiny objects, resulting in only a few pixels of feature representation for tiny objects in the final feature map. Weak feature responses are not enough for precise detection and location.

\textbf{Tiny objects lack network attention.} The feature responses of tiny objects are relatively weak and susceptible to interference, requiring extra attention. However, FPN applies the same processing method to features at each layer: $1\times1$ convolutions to reduce the number of channels, to-down features fusion and $3\times3$ convolutions to integrate outputs, without providing special treatment for tiny objects features.

\textbf{FPN lacks spatial perception ability.} By merging upper and lower layer features in a pixel-by-pixel addition manner, FPN lacks spatial perception ability around tiny targets. The pixel offset caused by recursive upsampling in FPN results in misalignment of tiny target features between upper and lower layers. Therefore, it is necessary to enhance the spatial perception ability of FPN and enrich the contextual information of tiny target features.

In fact, many traditional methods for detecting tiny objects are based on frequency domain techniques \cite{fre1, fre2}. For example, wavelet transform can be used to separate the original image into low-frequency and high-frequency components, and then thresholding can be applied to the high-frequency components to detect tiny objects \cite{fre3}. Because low-frequency components represent the overall contours and large, smooth areas of the image, while tiny objects typically appear as details and edges, which correspond to high-frequency components in the image's frequency spectrum. By filtering out the low-frequency components, tiny objects can be highlighted. This can be achieved by using discrete cosine transform (DCT) and a high-pass filter. The effectiveness of this filtering can be quantified using the Signal to Clutter Ratio (SCR), which reflects the degree of discrimination between the target and its surrounding background in the local image region. A higher SCR value generally indicates a more salient target. The SCR is mathematically defined as: 
\begin{equation} 
SCR = \frac{|\mu_{t} - \mu_{b}|}{\sigma_b} 
\label{eq0} 
\end{equation} 
where $\mu_{t}$ is the average gray value of the target region (a local image region of $40\times40$ pixels centered around the target in this work), and $\mu_{b}$ and $\sigma_{b}$ are the average and standard deviation of the gray value in the target's neighborhood, respectively. Figure \ref{fig1} shows the target regions, their 3D surface representations, and the SCR values after filtering out low-frequency components across different ranges. The results indicate that moderate filtering of low-frequency components can significantly improve SCR, making tiny objects more prominent.


Inspired by the above, we firstly design a high frequency perception module (HFP) to enhance the features of tiny objects by filtering out low-frequency components in the FPN feature maps. HFP begins with a predefined high-pass filter to extract the high-frequency response of the input feature map. This high-frequency response is used to refine the original feature map through both channel and spatial branches. The channel path dynamically allocates weights to each channel of the original feature map based on the high-frequency response, highlighting channels with more tiny object features. The spatial path allocates weights to each pixel in the original feature map based on the high-frequency response, generating a spatial mask that directs attention on the areas with tiny object features. Secondly, we design a spatial dependency perception module (SDP) operated between adjacent upper and lower feature maps. SDP calculates similarity at the pixel level and learns the spatial dependencies between adjacent pixel in the upper and lower feature maps, enriching tiny object features by leveraging valuable spatial dependencies.

Experiments show that using ResNet50 as the backbone, Faster R-CNN based on HS-FPN achieves an average precision ($AP$) of 20.3 on the AI-TOD dataset, compared to 18.3 with FPN. Similarly, Cascade R-CNN based on HS-FPN improves $AP$ from 20.2 to 23.6. Furthermore, HS-FPN has a similar overall structure to FPN and can easily embedded into any model that requires FPN.

The main contributions of this paper are as follows:
\begin{itemize}
\item We reveal three issues faced by current FPN-based object detection models in detecting tiny objects.
\item We propose a novel feature pyramid network, HS-FPN, which addresses these issues mentioned above through high frequency perception module and spatial dependency perception module.
\item We integrate HS-FPN into various detection models to replace FPN, and evaluate its performance on multiple tiny object detection datasets. Experiments indicate that HS-FPN significantly improves performance compared to FPN.
\end{itemize}

\begin{figure}
\centering
\includegraphics[width=0.85\columnwidth]{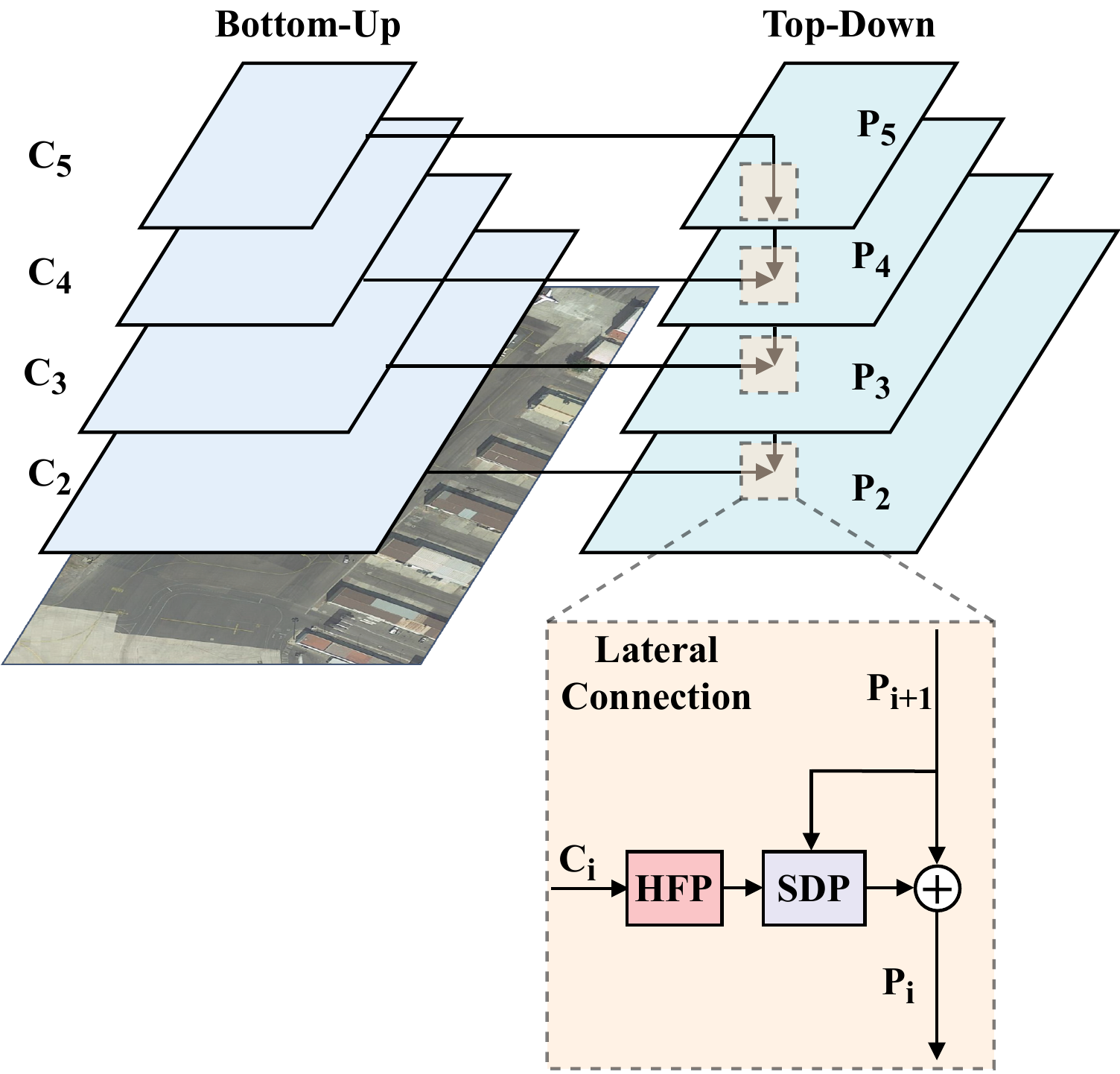} 
\caption{The overall structure of HS-FPN and the details of the lateral connection. HFP means high frequency perception module while SDP means spatial dependency perception module. For better visualization, convolution layers are omitted. Please note that SDP dose not appear in ${P_{5}}$.}
\label{fig2}
\end{figure}

\section{Related Work}
\subsection{Tiny Object Detection}
Most of the research on TOD focuses on three main aspects: benchmark construction, better label assignment strategy, and multi-scale learning. 

\textbf{Benchmark construction.} Generally used datasets like COCO \cite{coco} lack sufficient tiny object instances. To address this, new datasets such as AI-TOD \cite{aitod}, SODA \cite{soda}, and AI-TOD-v2 \cite{aitod-v2} have been developed, including more tiny object samples and effective evaluation metrics, advancing detection techniques.

\textbf{Better label assignment strategy.} The Intersection over Union (IoU) metric, commonly used for matching predicted boxes with ground truth, is highly sensitive to location deviations in tiny objects, significantly compromising label assignment quality in anchor-based detectors. There have already been some improvements to the IoU metric, methods such as Dot Distance \cite{dot-distance}, NWD \cite{aitod-v2}, and RFLA \cite{rfla}, have demonstrated effectiveness in TOD.

\textbf{Multi-scale learning.} As a classic multi-scale learning method, FPNs \cite{fpn, pafpn, cefpn} effectively integrates high-level semantics from deeper feature maps with detailed information from shallower ones through a top-down structure and lateral connections, significantly improving tiny object detection performance. 

\subsection{Learning in Frequency Domain}
Frequency domain analysis has long been a crucial tool in image processing. With the breakthroughs in deep learning, researchers have increasingly integrated frequency domain information into deep learning methods \cite{frequency-learning}. For instance, \citeauthor{gfnet} \shortcite{gfnet} used 2D discrete Fourier transform and a set of learnable frequency filters to enable frequency domain interactions between tokens, learning long-term spatial dependencies in the frequency domain with log-linear complexity. \citeauthor{fcanet} \shortcite{fcanet} mathematically demonstrated that the commonly used Global Average Pooling is equivalent to the lowest frequency component in the DCT and assigned weights to each channel of the feature map based on DCT. While these studies were not all specifically focused on tiny object detection, they have inspired our research.

\section{Method}
In this section, we detail the HS-FPN framework and its components. As shown in Figure \ref{fig2}, the structure of HS-FPN is similar to FPN, collecting four feature maps from the backbone network and reducing their channel dimensions to 256 using $1\times1$ convolutions. These reduced-channel features are named as \{${C_{2}, C_{3}, C_{4}, C_{5}}$\} which have strides of \{4, 8, 16, 32\} pixels relative to the input image. The feature pyramid \{${P_{2}, P_{3}, P_{4}, P_{5}}$\} is generated by the top-down pathway in HS-FPN. Each lateral connection in HS-FPN contains two modules: high frequency perception module (HFP) and spatial dependency perception (SDP) module. HFP generates high frequency responses to enrich the information of tiny objects in $C_{i}$ while SDP takes \{$C_{i}, P_{i+1}$\} as inputs using pixel-level cross attention mechanism to learn the relationships between pixels in \{$C_{i}, P_{i+1}$\} and then using valuable spatial dependencies to enrich the features of tiny targets. Please note that all laterals of HS-FPN contain the HFP module, while only \{${P_{2}, P_{3}, P_{4}}$\} layers contain the SDP module. Finally, the output features of HS-FPN at each layer are obtained from \{${P_{2}, P_{3}, P_{4}, P_{5}}$\} through separate $3\times3$ convolutions for subsequent detection tasks.

\begin{figure*}
    \centering
    \includegraphics[width=0.85\textwidth]{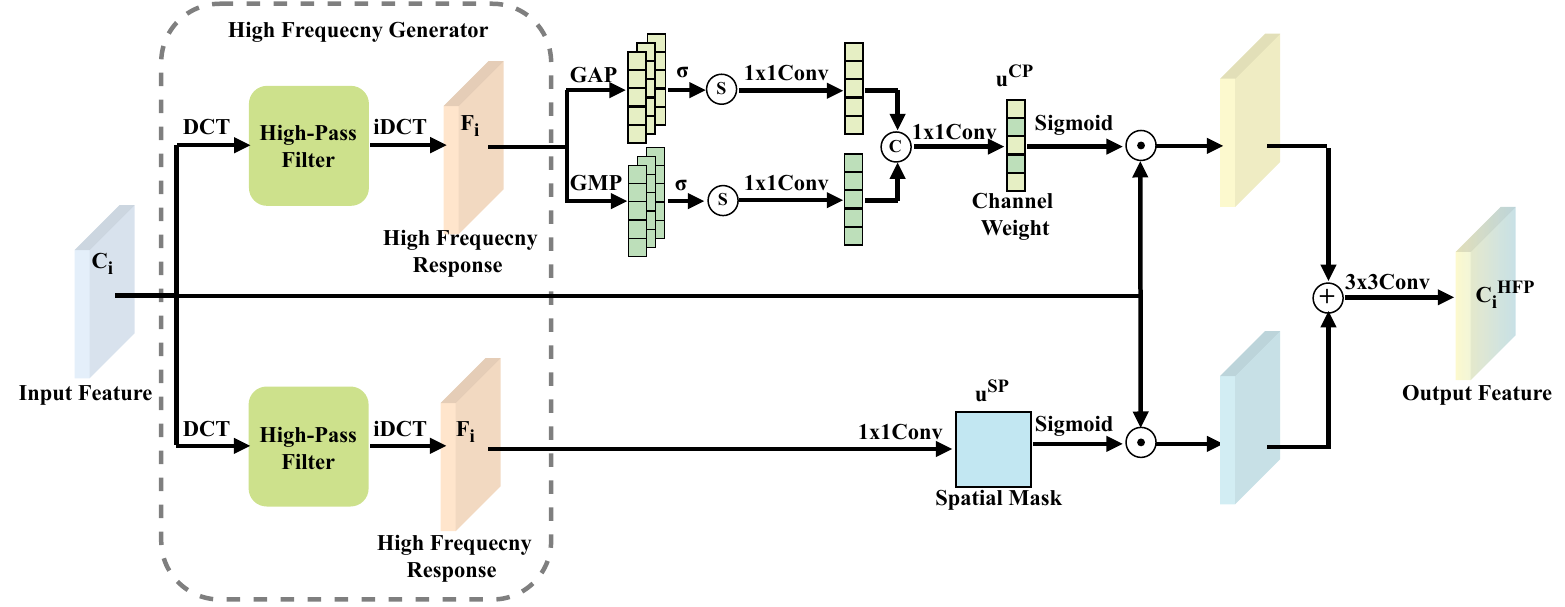}
    \caption{The design ideas of HFP, which consists of a high frequency feature generator, a channel path (CP), and a spatial path (SP). Please note that the high frequency feature extractor within the dashed box is implemented only in the \{${P_{2}, P_{3}}$\} and the two high frequency features in the diagram are identical. \circled[3]{·} means Hadamard product, \circled[1]{c} means feature cognate and \circled[2]{s} denotes pixel-by-pixel summation. $\sigma$ refers ReLU function.}
    \label{fig3}
\end{figure*}

\subsection{High Frequency Perception Module}
To address the issues mentioned in Section \ref{introduction} that tiny objects have limited usable features and FPN does not pay extra attention to tiny objects, we propose the high frequency perception module (HFP). As shown in Figure \ref{fig3}, HFP consists of a high-frequency feature generator, a channel path (CP), and a spatial path (SP). Assuming the input feature $C_{i}$ has a size of $R^{(C\times{H_{i}}\times{W_{i}})}$, HFP first applies a high-pass filter to $C_{i}$ to obtain the high frequency responses $F_{i}$ which has the same size of $C_{i}$. Then $F_{i}$ is used respectively as input for CP and SP to generate channel attention weights $u^{CP} \in R^{(C\times{1}\times{1})}$ and spatial attention weights $u^{SP} \in R^{(1\times{H_{i}}\times{W_{i}})}$. These weights are broadcast along the spatial and channel dimensions, respectively, and element-wise multiplied with $C_{i}$. Finally, the results from CP and SP are added and then passed through a $3\times3$ convolution to produce the output of HFP, denoted as ${C_{i}}^{HFP}$. 



\textbf{Channel Path.} 
It is undeniable that each channel of $C_{i}$ contributes differently to the representation of tiny objects. Therefore, we aim to identify the channels within $C_{i}$ that contain more tiny object features and assign them distinct weights. General channel attention \cite{scenet, cbam} is typically calculated across the entire feature map. However, since tiny object features occupy a relatively small proportion of the original feature map, the channel weights computed across the whole map are easily influenced by low-frequency homogeneous background. By applying a high-pass filter to $C_{i}$, most low-frequency components and homogeneous background are filtered out, increasing the proportion of tiny object features in the high-frequency responses $F_{i}$. As a result, calculating channel attention at $F_{i}$ becomes more accurate. 

To more effectively extract information from each channel, we chose to compress the spatial dimensions of $F_{i}$, ensuring that each channel of $F_{i}$ is representative. Global Average Pooling (GAP) can integrate the spatial information of the feature map to obtain an overall representation, while Global Max Pooling (GMP) captures the maximum activation value in each channel. Therefore, we use both GAP and GMP to process $F_{i}$, integrating its feature information to facilitate the subsequent calculation of each channel's importance to tiny objects. Firstly, as shown in Figure \ref{fig3}, we perform a GAP and a GMP on $F_{i}$ separately, resulting in $F_{i}^{GAP}$ and $F_{i}^{GMP}$. It is important to note that we directly pool $F_{i}$ into $R^{(C\times{k}\times{k})}$ instead of pooling it to the same size of $u^{CP}$, as that results in a significant loss of information. In this paper, $k$ is generally set to 16, as when k is too small, key information is lost in both GAP and GMP; when k is too large, the effectiveness of both operations diminishes. Next, we apply the ReLU function to both $F_{i}^{GAP}$ and $F_{i}^{GMP}$ to retain positive values, and then sum across each channel to generate two one-dimensional feature vectors. These vectors are then passed through separate $1\times1$ group convolutions to generate different channel scores. These scores are then concatenated and passed through another $1\times1$ group convolution, producing the final channel attention weights $u^{CP}$. 

\textbf{Spatial Path.}
The SP of HFP is designed to help the model better focus on tiny object features within the spatial range, achieving an effect similar to self-attention \cite{attentionisallyouneed}. However, self-attention relies on similarity calculations between feature pixels to emphasize important information. Given that tiny object features are weak and FPN's shallow feature maps contain noise, self-attention may struggle to effectively highlight tiny object features and could even amplify noise and background responses. With the help of the high-pass filter, the low-frequency homogeneous background in $F_{i}$ is filtered out and tiny object features are enhanced.
Therefore, we use $F_{i}$ as a spatial mask. Specifically, a $1\times1$ convolution aggregates the channel-wise information of $F_{i}$ to generate the spatial attention mask $u^{SP}$.

\begin{figure}
    \centering
    \includegraphics[width=0.5\columnwidth]{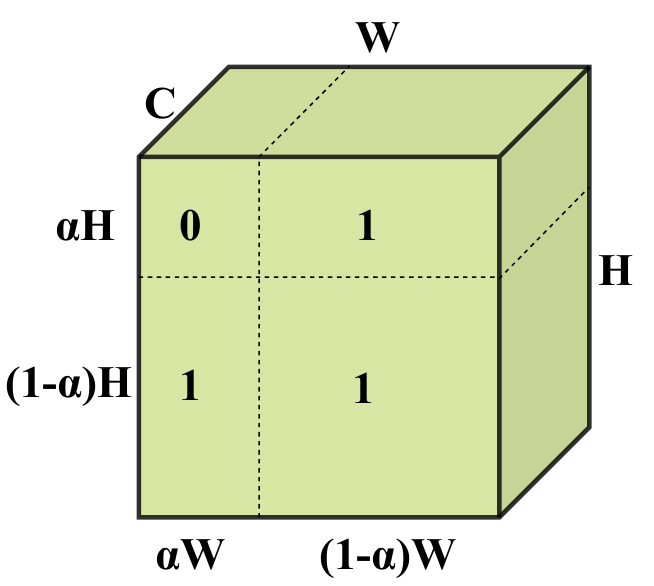} 
    \caption{Illustration of High-Pass Filter.}
    \label{fig4}
\end{figure}

\textbf{High-Pass Filter.} 
To highlight tiny object features in the input feature map, we designed a high-pass filter in HFP, as shown in Figure \ref{fig4}. The filter $z$ has the same size of $F_{i}$, assumed to be ${R}^{(C\times{H_{i}}\times{W_{i}})}$. Since the low-frequency components of the DCT spectrum are concentrated in the top-left corner and the high-frequency components are distributed towards the bottom-right corner, we introduce a hyper-parameter $\alpha$ to control the frequency band of the filter. Specifically, the values of filter $z$ are determined as follows:
\begin{equation}
    z(u,v) = \begin{cases} 
    0 &u<\alpha{H}, v<\alpha{W}\\
    1 &else 
    \end{cases}, \alpha \in [0, 1]
    \label{eq2}
\end{equation}
When $\alpha=0$, all values of filter $z$ are 1, making $F_{i}$ identical to $C_{i}$. When $\alpha=1$, all values of filter $z$ are 0, blocking all frequencies. The high-pass filter $z$ only works when $\alpha$ is between 0 and 1, filtering out low-frequency components from $C_{i}$ to generate high-frequency responses $F_{i}$.

\begin{figure}[t]
    \centering
    \includegraphics[width=0.85\columnwidth]{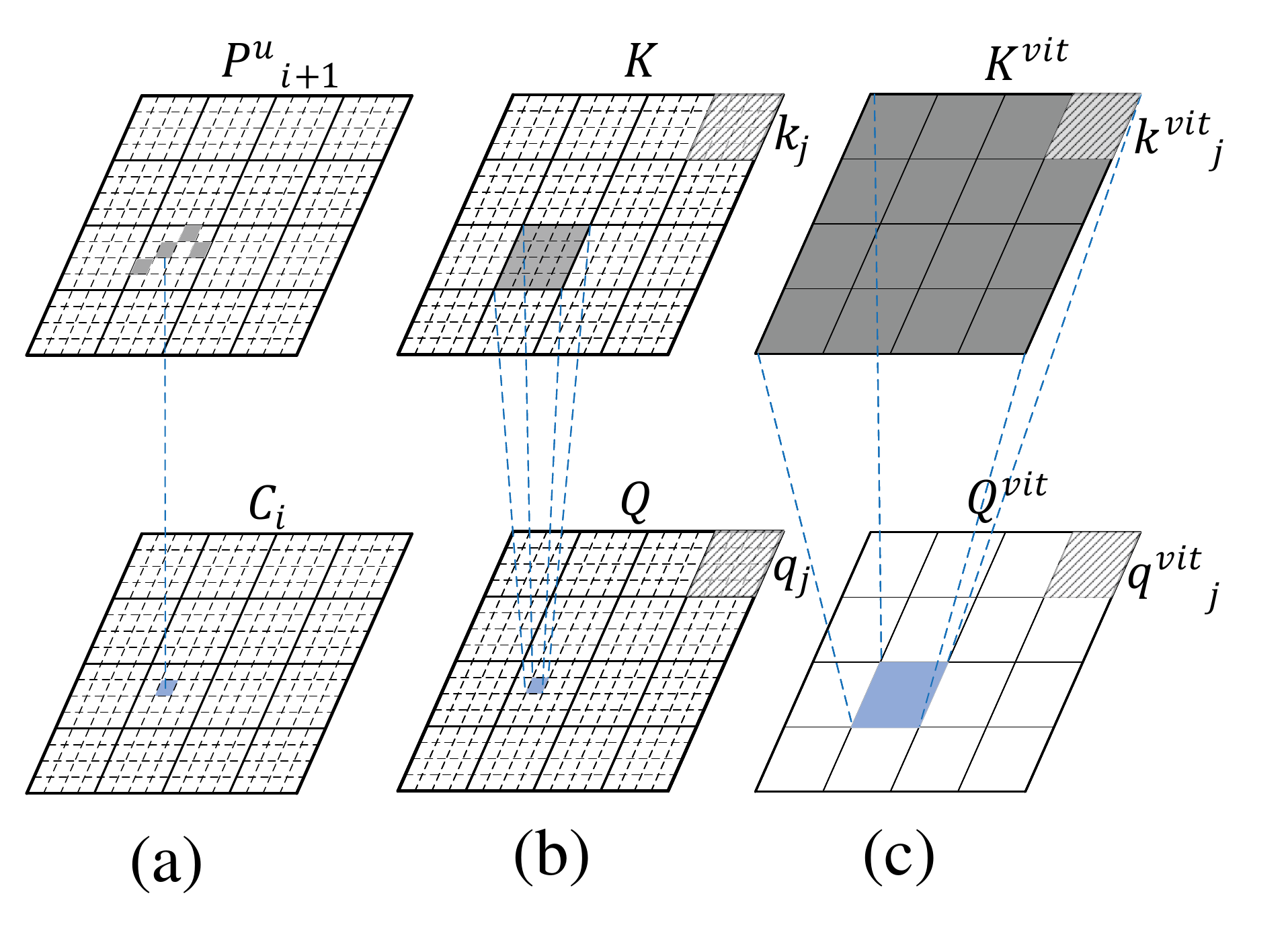} 
    \caption{Computing details between different attention methods. (a) Lack of spatial dependency in FPN; (b) Attention between pixels used in SDP. (c) Attention between feature blocks used in ViT.}
    \label{fig5}
\end{figure}

\begin{figure}
    \centering
    \includegraphics[width=0.95\columnwidth]{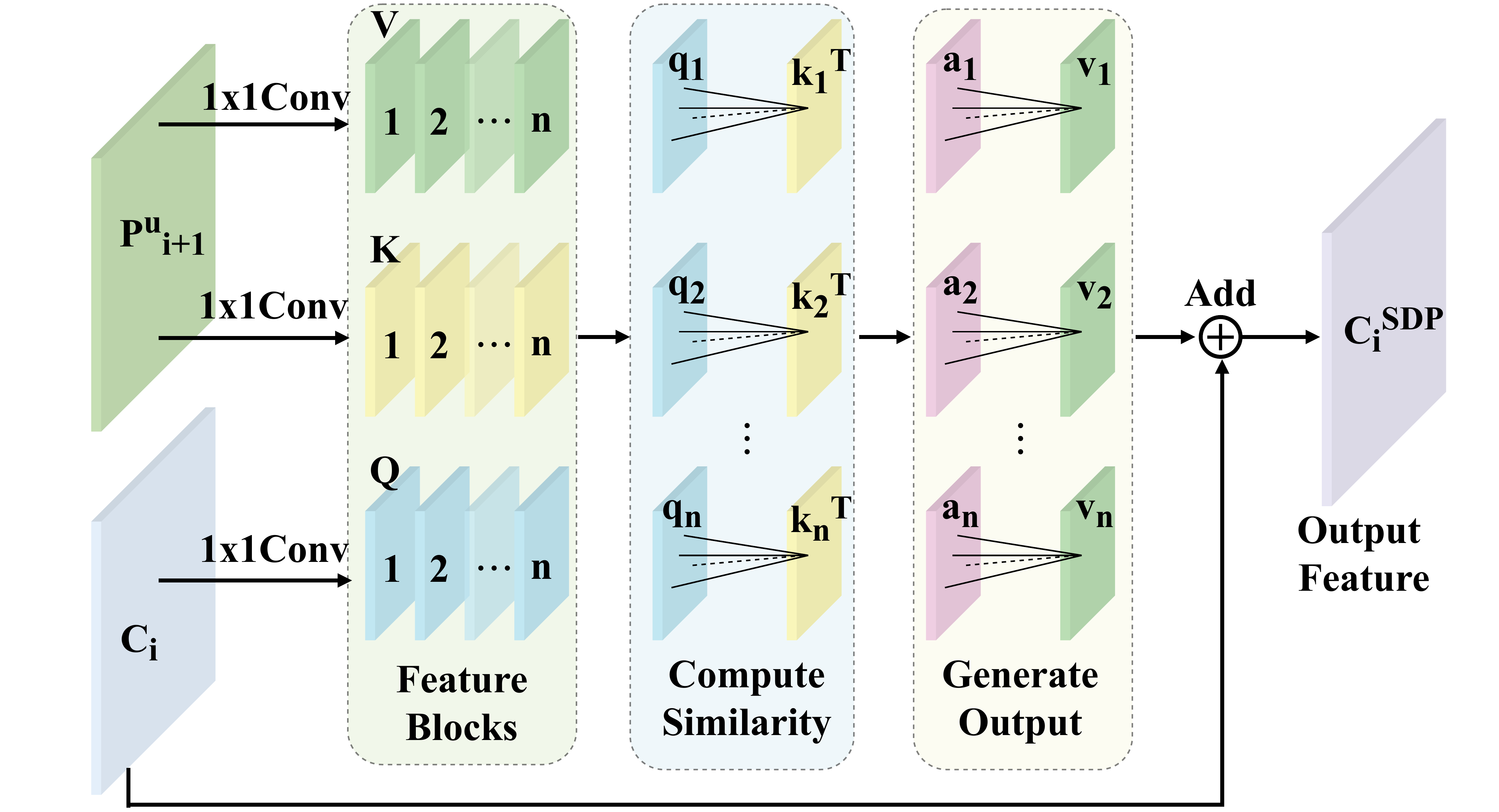} 
    \caption{An over review of SDP. $P^{u}_{i+1}$ is upper feature and has already been upsampled.}
    \label{fig6}
\end{figure}

\subsection{Spatial Dependency Perception Module}
By merging upper and lower layer features in a pixel-by-pixel addition manner, FPN lacks spatial perception ability around tiny targets, as shown in Figure \ref{fig5} (a). However, enhancing spatial perception is crucial. For instance, the pixel offset caused by recursive upsampling in FPN results in misalignment between features \{${C_{i}, P_{i+1}}$\}, which greatly affects the accuracy of tiny object features. Inspired by the attention mechanism in Vision Transformer (ViT) \cite{vit}, we designed the spatial dependency perception module (SDP) to capture the mapping between pixels in \{${C_{i}, P_{i+1}}$\}. The attention mechanism, proficient at capturing long-range dependencies, enables pixels from lower-level features $C_{i}$ to interact not only with pixels at the same spatial location in $P_{i+1}$, but also with a wider range of pixels. Ultimately, we leverage the valuable spatial dependencies captured before to enrich the features of tiny objects. 

As shown in Figure \ref{fig6}, we first upsample $P_{i+1}$ to match the size of $C_{i}$, and name the upsampled feature $P^{u}_{i+1}$. Subsequently, we compute the Query (Q), Key (K), and Value (V) matrices through three separate $1\times1$ convolutions, where Q is derived from $C_{i}$ and K, V are derived from $P^{u}_{i+1}$. Next, we divide the \{Q, K, V\} matrices into multiple feature blocks. For simplicity, let's assume the dimensions of \{Q, K, V\} are $R^{(C\times{H_{i}}\times{W_{i}})}$. Since the detection framework doesn't constrain input image size, the feature block size must be adjusted based on the actual input image to avoid issues with non-divisible dimensions. Please note that the input feature maps in the feature pyramid \{${C_{2}, C_{3}, C_{4}, C_{5}}$\} are multiples of each other in size, so making the feature block have the same size with $C_{5}$ could ensure a perfect division. Based on this assumption, we partition \{Q, K, V\} into feature blocks and reshape them into $R^{(n\times({H_{5}}\times{W_{5})\times{C}})}$, where $n=\frac{H_{i}}{H_{5}}\times\frac{W_{i}}{W_{5}}$, $C$ is the number of channels in $C_{i}$, ${H_{5}}\times{W_{5}}$ is the resolution of $C_{5}$. It can be explained as: $C_{i}$ can be divided into $n$ feature blocks with each block containing ${H_{5}}\times{W_{5}}$ feature points, and each point is represented as a $C$ dimensional column vector. Then we extract corresponding feature blocks $q_{j}$ and 
$k_{j}$ ($1\leq{j}\leq{n}, q_{j}, k_{j} \in R^{((H_{5}\times{W_{5}})\times{C})}$) from \{Q, K\} and calculate their similarity and obtain the similarity matrix $a_{j} \in R^{((H_{5}\times{W_{5}})\times{(H_{5}\times{W_{5}})})}$. Finally, each feature point in the corresponding block $v_{j}$ from V is weighted by $a_{j}$ to generate feature blocks rich in spatial dependencies and high-level semantics.
\begin{equation}
    a_{j}=Softmax(\frac{q_{j}\times{{k_{j}}^{T}}}{\sqrt{C}}), 1\leq{j}\leq{n}
    \label{eq3}
\end{equation}
After applying the cross-attention mechanism to each feature block of Q, the blocks are aggregated according to their spatial locations to form a new feature rich in spatial dependencies, which is then added to $C_{i}$. This process effectively enriches the feature content of tiny objects and infuses them with the rich semantics of high-level features.

Please note that the attention calculation process in SDP differs from that in ViT. First, ViT reshapes Q into $R^{(n\times({H_{5}}\times{W_{5}\times{C})})}$, which means Q consists of $n$ feature points with each being a vector of dimensions $({H_{5}}\times{W_{5}\times{C}})$. Second, there is only one similarity matrix of size $R^{(n\times{n})}$ in ViT instead of $n$ similarity matrices in SDP. In summary, SDP computes cross-attention between pixels within feature blocks, while ViT computes attention across feature blocks. These differences can be intuitively understood from Figure \ref{fig5} (b, c).

\section{Experiments and Analysis}
\subsection{Datasets and Metrics}
Experiments are performed on two TOD datasets. The main experiments are conducted on the challenging AI-TOD \cite{aitod}, which provides a total of 700,621 instances in 8 categories in 28,036 aerial images. The average absolute instance size of AI-TOD is 12.8 pixels. Furthermore, we build a subset consisting of 10 categories named $DOTA_{mini10}$ from DOTA \cite{dota} and test our method on it. This subset contains 410,305 instances, with $17.3\%$ of the instances smaller than 16 pixels and $38.5\%$ smaller than 32 pixels. All metrics used in the experiments follow the AI-TOD benchmark.

\subsection{Experiments Settings}
All of our experiments are implemented based on MMDetection \cite{mmdetection} and PyTorch \cite{pytorch}. We train and evaluate the detectors on two NVIDIA 3080 Ti GPUs (one image per GPU). All detectors are trained using the Stochastic Gradient Descent (SGD) optimizer for 12 epochs with a momentum of 0.9 and weight decay of 0.0001. The initial learning rate of two-stage models like Faster R-CNN \cite{faster-rcnn} and one-stage models like RetinaNet \cite{retina} is set to 0.005 and 0.001 respectively, and decays at the $8^{th}$ and $11^{th}$ epochs. Besides, the basic anchor size is set to 2 for all anchor-based models.

\begin{table}
  \centering
  \setlength{\tabcolsep}{0.9mm}{
    \begin{tabular}{ccc|ccc|cccc}
    \toprule       
\multicolumn{2}{c|}{HFP} & \multirow{2}{*}[-1ex]{SDP} & \multirow{2}{*}[-1ex]{$AP$} & \multirow{2}{*}[-1ex]{${AP_{50}}$} & \multirow{2}{*}[-1ex]{${AP_{75}}$} & \multirow{2}{*}[-1ex]{${AP_{vt}}$} & \multirow{2}{*}[-1ex]{${AP_{t}}$} & \multirow{2}{*}[-1ex]{${AP_{s}}$} & \multirow{2}{*}[-1ex]{${AP_{m}}$} \\
\cmidrule{1-2}          
\multicolumn{1}{c|}{CP} & \multicolumn{1}{c|}{SP} &       &       &       &       &       &       &       &  \\
    \midrule
\textcolor[RGB]{190,190,190}{\ding{55}}    & \textcolor[RGB]{190,190,190}{\ding{55}}     & \textcolor[RGB]{190,190,190}{\ding{55}}     & 20.2  & 47.4  & 13.8  & 9.9   & 21.3  & 24.1  & 30.3  \\
\ding{51}     & \textcolor[RGB]{190,190,190}{\ding{55}}     & \textcolor[RGB]{190,190,190}{\ding{55}}     & 21.6      & 49.2      & 15.2      & 9.6      & 23.0      & 25.7      & 30.8 \\
\textcolor[RGB]{190,190,190}{\ding{55}}     & \ding{51}     & \textcolor[RGB]{190,190,190}{\ding{55}}     & 21.6      & 49.6      & 15.3      & 10.0      & 23.1      & 25.4      & \textbf{31.7} \\
\ding{51}    & \ding{51}     & \textcolor[RGB]{190,190,190}{\ding{55}}    & \underline{22.4}  & \underline{50.4}  & \underline{16.1}  & \textbf{12.1}  & \underline{23.9}  & \underline{26.1}  & 31.5  \\
\textcolor[RGB]{190,190,190}{\ding{55}}     & \textcolor[RGB]{190,190,190}{\ding{55}}     & \ding{51}     & 21.3  & 49.1  & 15.1  & 8.3   & 23.1  & 25.3  & 30.3  \\
\ding{51}     & \ding{51}     & \ding{51}     & \textbf{23.6}   & \textbf{52.3}  & \textbf{17.4}  & \underline{11.6}  & \textbf{25.2}  & \textbf{27.0}  & \underline{31.6}  \\
    \bottomrule
    \end{tabular}%
    }
      \caption{Effect of each component on AI-TOD test set. 
      }
  \label{tab1}%
\end{table}%

\begin{table}
  \centering
  \setlength{\tabcolsep}{1.5mm}{
    \begin{tabular}{c|ccc|cccc}
    \toprule
 $\alpha$  &   $AP$ & ${AP_{50}}$ & ${AP_{75}}$ & ${AP_{vt}}$ & ${AP_{t}}$ & ${AP_{s}}$ & ${AP_{m}}$ \\
    \midrule
    0.00  & 21.7  & 49.3  & 15.5  & 9.7   & 23.2  & \textbf{26.4}  & 29.9  \\
    0.25  & \textbf{22.4}  & \underline{50.4}  & \textbf{16.1}  & \textbf{12.1}  & \textbf{23.9}  & \underline{26.1}  & \textbf{31.5}  \\
    0.50  & \underline{22.1}  & 49.9  & \underline{15.9}  & 10.3  & \underline{23.6}  & 25.8  & \underline{31.3}  \\
    0.75  & 21.9       & \textbf{50.7}       & 15.4      & \underline{10.7}      & 23.3      & 25.9      & 31.0 \\
    1.00  & 21.5  & 49.0  & 15.2  & 10.2  & 22.7  & 25.7  & 30.1  \\
    \bottomrule
    \end{tabular}%
    }
      \caption{Optimal studies of $\alpha$ on AI-TOD test set.}
  \label{tab2}%
\end{table}%

\begin{table*}[t]
  \centering
  \setlength{\tabcolsep}{1mm}{
    \begin{tabular}{l|l|*{3}{c}|*{4}{c}}
    \toprule
   Method & Backbone & $AP$ & ${AP_{50}}$ & ${AP_{75}}$ & ${AP_{vt}}$ & ${AP_{t}}$ & ${AP_{s}}$ & ${AP_{m}}$ \\
    \midrule
    ATSS \cite{atss}  & R-50 + FPN & 15.5  & 38.2  & 10.1  & 4.0   & 14.5  & 21.5  & 31.9  \\
    MENet \cite{menet} & R-50 + FPN & 20.4  & 50.0  & 12.9   & 8.9   & 21.4  & 23.2  & 31.0  \\
    RetinaNet* \cite{retina} & R-50 + FPN & 13.7  & 31.5  & 9.4   & 5.8   & 15.7  & 15.0  & 16.6  \\
    Cascade RPN \cite{cascadeRPN} & R-50 + FPN & 18.1  & 44.9  & 11.2  & 7.1   & 18.8  & 22.3  & 27.9  \\
    Grid R-CNN \cite{grid} & R-50 + FPN & 18.4  & 38.7  & 14.6  & 12.1  & 19.0  & 21.8  & 25.5  \\
    Faster R-CNN \cite{faster-rcnn} & R-50 + FPN & 18.3  & 44.7  & 11.3  & 9.4   & 19.4  & 22.4  & 27.0  \\
    Faster R-CNN w/NWD \cite{aitod-v2} & R-50 + FPN & 19.4  & 47.3  & 12.5  & 8.0   & 20.4  & 22.7  & 28.6  \\
    Faster R-CNN w/RFLA \cite{rfla} & R-50 + FPN & 20.9  & 50.8  & 13.0  & 8.1   & 21.2  & 25.9  & 32.7  \\
    Faster R-CNN \cite{mobile} & MobileNetV2  + FPN & 15.6  & 39.2  & 9.2   & 9.8   & 16.9  & 18.7  & 20.5  \\
    Faster R-CNN & R-101 + FPN & 18.8  & 45.8  & 11.6  & 11.1  & 19.7  & 22.9  & 27.9  \\
    Cascade R-CNN & R-50 + FPN & 20.2  & 47.4  & 13.8  & 9.9  & 21.3  & 24.1  & 30.3  \\
    Cascade R-CNN w/NWD & R-50 + FPN & 21.4  & 49.1  & 15.1  & 9.4   & 22.4  & 25.1  & 31.5  \\
    Cascade R-CNN w/RFLA & R-50 + FPN & 21.9  & 50.6  & 15.9  & 8.3   & 21.9  & 26.9  & 34.7  \\
    DetectoRS \cite{detectors} & R-50 w/SAC + RFP & 22.8  & 51.4  & 16.9  & 10.7  & 23.7  & 27.4  & 32.3  \\
    DetectoRS w/RFLA & R-50 w/SAC + RFP & 23.9  & 54.6  & 17.8  & 9.8  & 24.0  & \underline{28.9}  & \textbf{37.4}  \\
    \midrule
    RetinaNet* & R-50 + HS-FPN & 15.0[+1.3]  & 33.5  & 10.8  & 7.4   & 17.4  & 15.8  & 16.7  \\
    Cascade RPN & R-50 + HS-FPN & 19.3[+1.2]  & 46.3  & 12.7  & 10.5  & 20.4  & 23.9  & 30.1  \\
    Grid R-CNN & R-50 + HS-FPN & 19.3[+0.9]  & 40.6  & 15.7  & 11.3  & 21.0  & 21.6  & 26.6  \\
    Faster R-CNN & R-50 + HS-FPN & 20.3[+2.0]  & 48.8  & 13.3  & 11.6  & 22.0  & 25.5  & 27.8  \\
    Faster R-CNN w/NWD & R-50 + HS-FPN & 21.7[+2.3]  & 51.5  & 14.7  & 11.0  & 23.7  & 23.8  & 28.8  \\
    Faster R-CNN w/RFLA & R-50 + HS-FPN & 23.1[+2.2]  & 54.3  & 15.5  & 9.9   & 23.8  & 27.4  & 33.6  \\
    Faster R-CNN & MobileNetV2 + HS-FPN & 17.5[+1.9]  & 42.0  & 11.3  & 9.9   & 19.2  & 20.9  & 22.9  \\
    Faster R-CNN & R-101 + HS-FPN & 20.5[+1.7]  & 48.9  & 13.4  & 12.2  & 22.1  & 23.9  & 27.4  \\
    Cascade R-CNN & R-50 + HS-FPN & 23.6[\textbf{+3.4}]  & 52.3  & 17.4  & 11.6  & 25.2  & 27.0  & 31.6  \\
    Cascade R-CNN w/NWD & R-50 + HS-FPN & 24.3[\underline{+2.9}]  & \underline{54.8}  & 17.9  & \underline{12.7}  & 25.2  & 28.8  & 32.9  \\
    Cascade R-CNN w/RFLA & R-50 + HS-FPN & \underline{24.6}[+2.7]  & 54.6  & \underline{18.6}  & 9.2   & 25.0  & 28.6  & 35.9  \\
    DetectoRS & R-50 w/SAC + HS-FPN & 24.2[+1.4]  & 54.5  & 17.8  & \textbf{13.1} & \textbf{25.8} & 27.6  & 33.2  \\
    DetectoRS w/RFLA & R-50 w/SAC + HS-FPN & \textbf{25.1}[+1.2] & \textbf{55.7} & \textbf{19.1} & 12.1  & \underline{25.3}  & \textbf{29.9} & \underline{36.9} \\
    \bottomrule
    \end{tabular}%
    }
    \caption{Main results on AI-TOD. Models are trained on the train-val set and evaluated on the test set. Note that RetinaNet* means using P2-P6 of FPN. The relative improvements of $AP$ are shown in parenthesis. $\alpha$ is set to 0.25. "w/" means with.}  
  \label{tab3}%
\end{table*}

\begin{table}
  \centering
  \setlength{\tabcolsep}{0.5mm}{
    \begin{tabular}{l|*{2}{c}|*{3}{c}}
    \toprule
    Method & $AP$ & ${AP_{50}}$  & ${AP_{t}}$ & ${AP_{s}}$ & ${AP_{m}}$ \\
    \midrule
    RetinaNet* + FPN  & 38.2  & 68.1     & 13.7  & 31.8  & 44.9  \\
    Faster R-CNN + FPN  & 46.9  & 74.2    & 18.6  & 38.5  & 54.3  \\
    Cascade R-CNN + FPN & \underline{49.4}  & 74.2     & 18.1  & 40.0  & \underline{58.1}  \\
    \midrule
    RetinaNet* + HS-FPN  & 40.1  & 71.1      & 15.9  & 32.9  & 47.0  \\
    Faster R-CNN + HS-FPN & 48.4  & \underline{75.7}    & \underline{21.7}  & \underline{40.3}  & 55.9  \\
    Cascade R-CNN + HS-FPN & \textbf{50.9} & \textbf{76.6} & \textbf{22.2} & \textbf{41.8} & \textbf{59.0} \\
    \bottomrule
    \end{tabular}%
    }
  \caption{Results on $DOTA_{mini10}$. $\alpha$ is set to 0.25.}
  \label{tab4}%
\end{table}

\begin{table}
  \centering
    \setlength{\tabcolsep}{1mm}{
    \begin{tabular}{l|ccc|c|c}
    \toprule
    Method & $AP$ & $AP_{t}$ &$AP_{s}$   & FLOPs(G) & Params(M) \\
    \midrule
    Baseline (FPN) & 20.2 & 21.3  & 24.1 &162.34 &68.95 \\ 
    only CP & 21.6 & 23.0      & 25.7 & 193.67  & 71.34 \\
    only SP & 21.6  & 23.1      & 25.4 & 193.69  & 71.31 \\
    only HFP & \underline{22.4}  & \underline{23.9}  & \underline{26.1} & 193.69  & 71.34 \\
    only SDP & 21.3  & 23.1  & 25.3 & 169.32  & 69.35 \\
    HS-FPN (ALL) & \textbf{23.6}  & \textbf{25.2}  & \textbf{27.0} & 200.65  & 71.73 \\
    \bottomrule
    \end{tabular}%
    }
    \caption{FLOPs and parameters comparison for HS-FPN components.}
  \label{tab5}%
\end{table}%

\subsection{Ablation Study}
We analyze the effect of each proposed component of HS-FPN on AI-TOD test subset. Results are reported in Table \ref{tab1}. We progressively incorporate the channel path (CP) and spatial path (SP) of HFP and SDP into FPN. Baseline detector is Cascade R-CNN \cite{cascade} with ResNet50 \cite{resnet} as backbone.

\begin{figure}
    \centering
    \includegraphics[width=0.90\columnwidth]{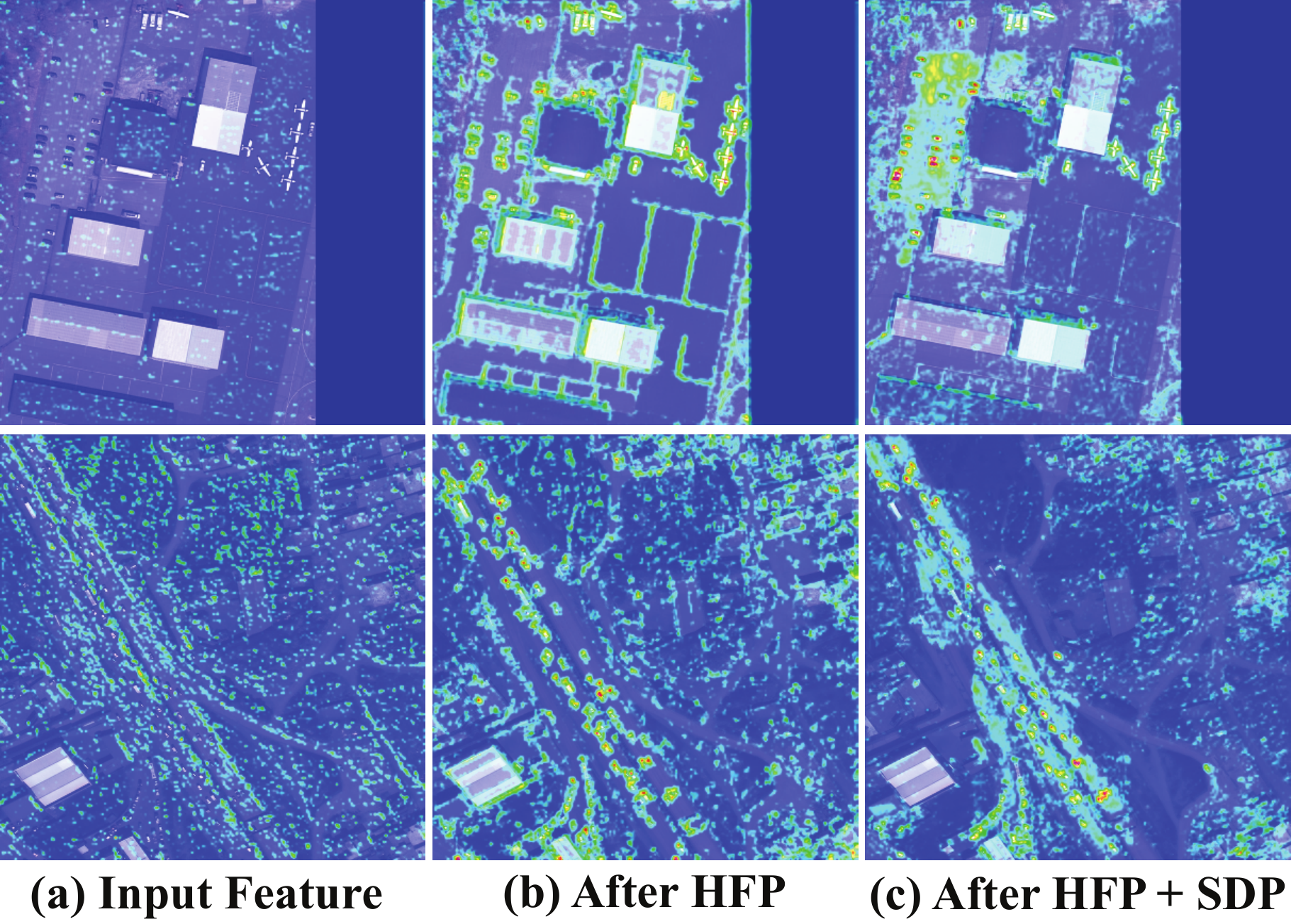} 
    \caption{Visualization of the feature at the $P_{2}$ level after the gradual addition of HFP and SDP in FPN.}
    \label{fig7}
\end{figure}

\textbf{The effect of High Frequency Perception Module.}
We implement HFP independently on each lateral layer of FPN. Results show that naively added HFP to FPN can bring 2.2 $AP$ points higher than baseline detectors, as well as $AP_{t}$ and $AP_{s}$ achieve 2.6 points and 2.0 points higher than baseline, respectively. This indicates that HFP has a positive effect on improving the performance of TOD. As shown in Figure \ref{fig7} (a, b), the tiny object features output by HFP are significantly enhanced compared to the input features. 

Furthermore, we also report the contribution of CP and SP within HFP. As shown in Table \ref{tab1}, when adding CP and SP individually to FPN, $AP$ both achieve 1.4 points improvement as well as $AP_{t}$ and $AP_{s}$ are also improved. Moreover, when CP and SP are combined in HFP, $AP$ is improved by 2.2 compared to FPN, and by 0.8 compared to using only SP or CP, demonstrating that the combination of CP and SP has a synergistic positive effect. 

In HFP, we use the parameter $\alpha$ to control the range of the high-pass filter. To investigate the impact of different filtering rates on tiny object detection, we conducted optimization experiments for $\alpha$. The results show that as $\alpha$ increases from 0 to 1, detection performance first improves and then declines, following a similar trend to the SCR depicted in Figure \ref{fig1}. This confirms the effectiveness of the HFP module and indicates that filtering out low-frequency information within an appropriate range can enhance the detection performance of tiny objects. Results are summarized in Table \ref{tab2}. Finally, in this paper, $\alpha$ is set to 0.25 for both AI-TOD and $DOTA_{mini10}$.

\textbf{The effect of Spatial Dependency Perception Module.}
We implement SDP on each lateral layer of FPN independently to evaluate its performance, resulting in an improvement of $AP$ by 1.1 points. When combined with HFP, the $AP$ increased by 3.4 points compared to FPN, 1.2 points over HFP alone, and 2.3 points over SDP alone, showing the benefits of their combination. As shown in Figure \ref{fig7} (c), the combination of HFP and SDP effectively enriches and enhances the features of tiny objects, while also suppressing some of the high-frequency noise amplified by HFP.

\subsection{Main Result}
We evaluate HS-FPN on the AI-TOD and $DOTA_{mini10}$ dataset, comparing its performance with the baseline model and other state-of-the-art models. The results are presented in Tables \ref{tab3} and \ref{tab4}. By simply replacing FPN with HS-FPN, nearly all baseline models show significant improvements in $AP$. Even with a more powerful backbone, like ResNet101, or a lightweight network like MobileNetV2, the $AP$ on Faster R-CNN still improved by 1.7 and 1.9 on AI-TOD, respectively. Additionally, using advanced methods like NWD and RLFA to optimize label assignment in anchor-based detectors further enhanced HS-FPN's performance, yielding notable $AP$ improvements in Faster R-CNN, Cascade R-CNN, and DetectoRS.

\subsection{Computational Costs}
Table \ref{tab5} demonstrates the computation increase of each component of HS-FPN within the entire TOD framework. The entire detection model is Cascade R-CNN with ResNet50 as the backbone and an input size of (800, 800). The Params and FLOPs for HFP mainly come from the $3\times3$ convolutions in Figure \ref{fig2}, which results in similar computational costs for CP and SP. Given the context of the entire detection framework, HS-FPN slightly increases computational cost but significantly improves performance compared to FPN.

\section{Conclusion}
In this paper, we propose a high frequency and spatial perception feature pyramid network (HS-FPN) to enhance the detection performance of tiny objects. Given the small scale and low-quality feature representation of tiny objects, we designed a high frequency perception module (HFP) to improve feature representation from the frequency domain. Additionally, we introduced a spatial dependency perception module (SDP) to capture the spatial dependencies between adjacent pixel points in the upper and lower feature maps, enriching the features of tiny objects. Extensive experiments on two tiny object detection datasets validate the effectiveness of our proposed method.

\textbf{Extended version.} Additional motivation and details are provided in the extended version\footnote{\url{https://arxiv.org/abs/2412.10116}}.

\section*{Acknowledgments}
This work was partly supported by the National Key Laboratory of Science and Technology on Multi-Spectral Information Processing.

\bigskip

\bibliography{aaai25}

\setcounter{secnumdepth}{2} 

\maketitle

\appendix
\section{Technical Appendix} 
\subsection{Motivation of HFP}
\begin{figure}[t]
    \centering
    \includegraphics[width=0.95\columnwidth]{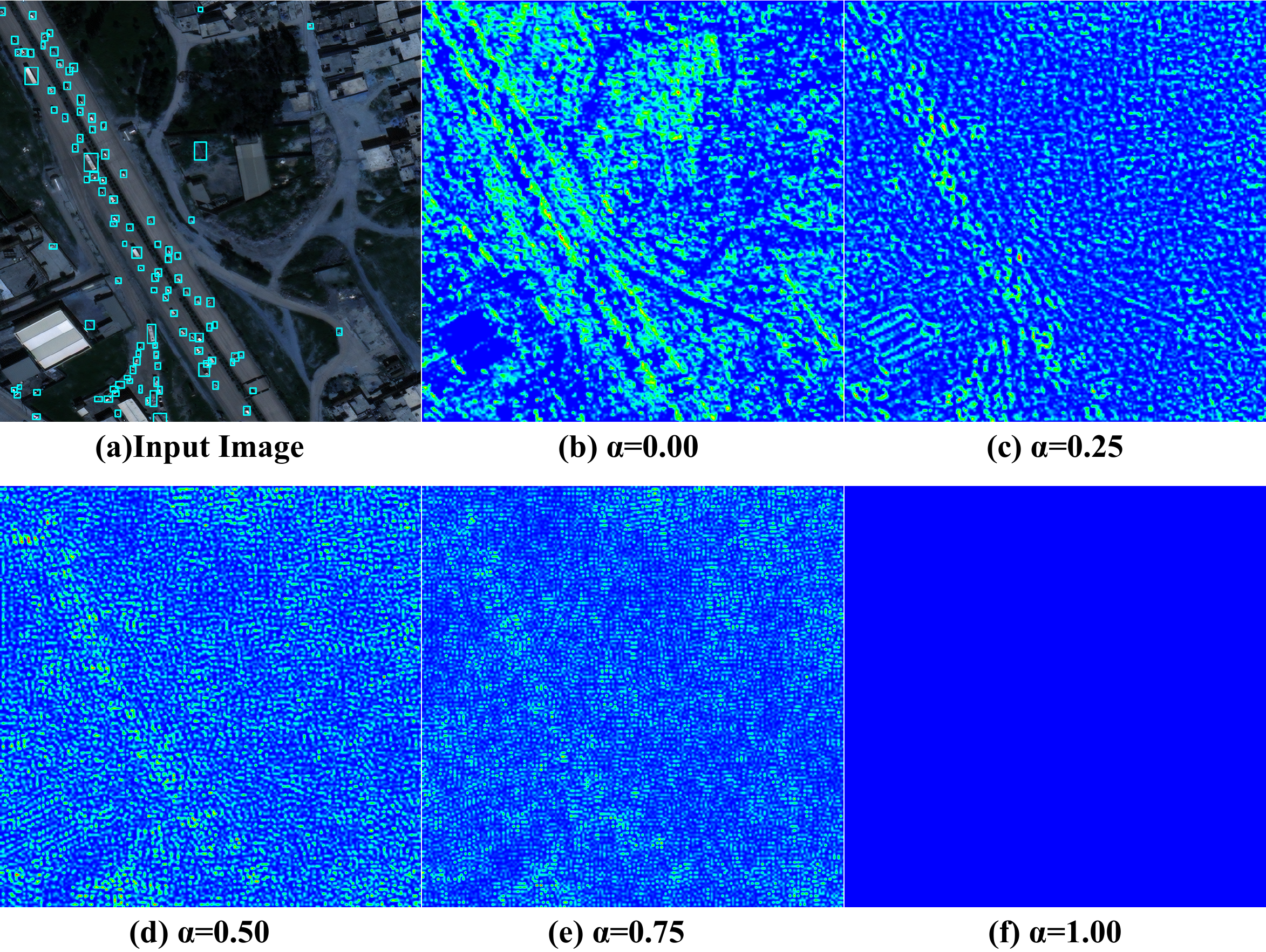} 
    \caption{Visualization of high-frequency responses $F_{2}$ generated under different $\alpha$.}
    \label{fig11}
\end{figure}
In this subsection, we discuss the motivation behind the High Frequency Perception (HFP) module and present the corresponding experimental results.

Figure \ref{fig11} presents the visualization of the high-frequency responses $F_{2}$ extracted from $C_{2}$ under different $\alpha$. It can be observed that when $\alpha$ is 0, all frequency bands can pass through filter $z$ making $F_{2}$ identical to the input feature $C_{2}$, which contains a large amount of low-frequency background information. As $\alpha$ gradually increases, the low-frequency components in $C_{2}$ are progressively filtered out by $z$, and the tiny object features in the generated high-frequency responses $F_{2}$ become increasingly prominent, with their proportion in $F_{2}$ significantly higher than in $C_{2}$. However, when $\alpha$ continues to increase beyond an optimal value, even the tiny object features in $C_{2}$ start to be filtered out by filter $z$, which manifests as the high-frequency noise becoming more pronounced in sub-figures (c-e), while the target features gradually weaken. When $\alpha$ reaches its maximum value of 1, $z$ completely blocks all frequency bands, resulting in the high-frequency response $F_{2}$ becoming an empty matrix, as shown in sub-figure (f).

\begin{figure}[t]
    \centering
    \includegraphics[width=0.95\columnwidth]{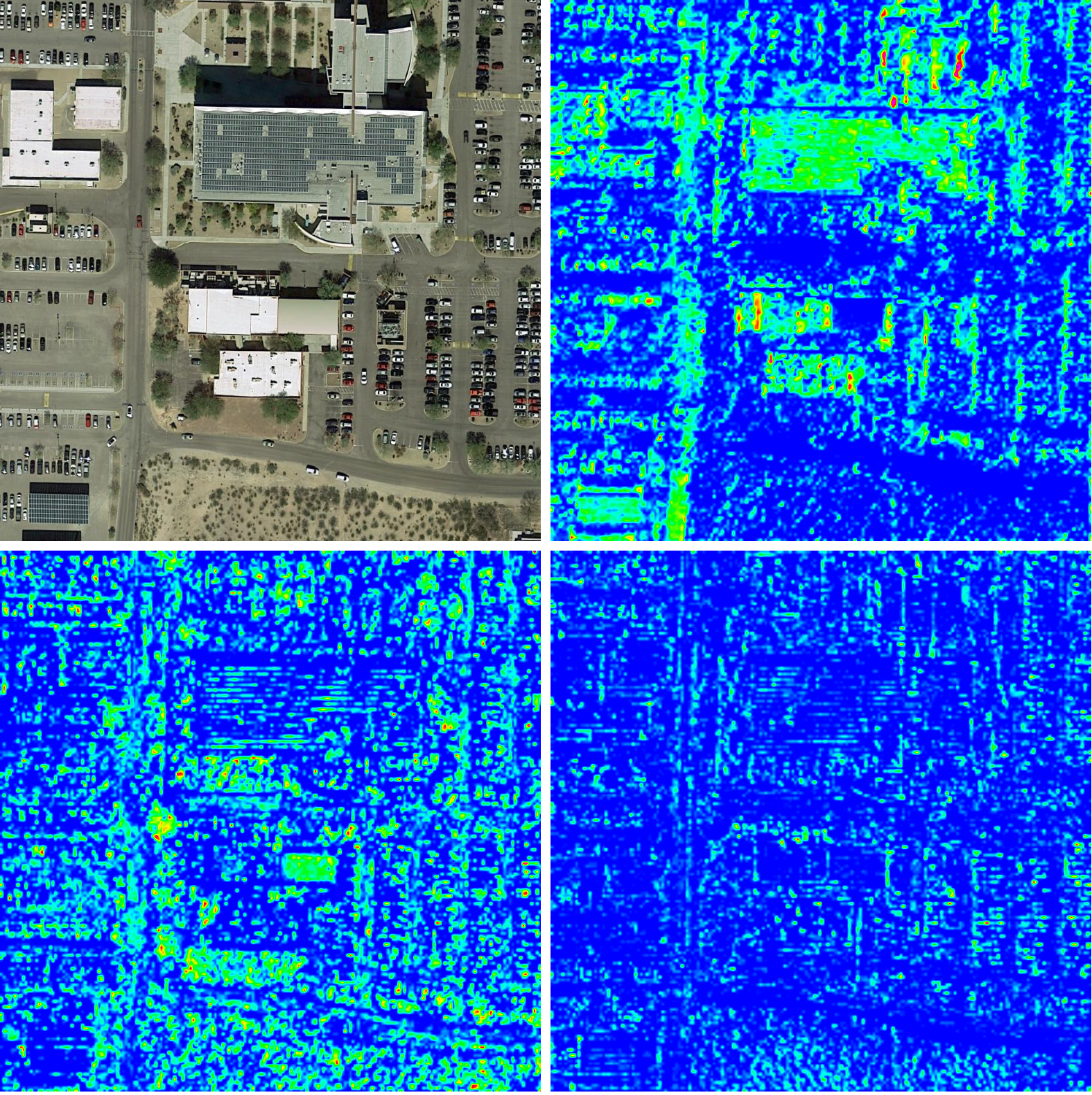} 
    \caption{Input image and partial channel feature visualization of $C_{2}$.}
    \label{fig12}
\end{figure}

It can be observed from Figure \ref{fig12}, due to the different parameters of convolutional kernels in each channel, the extracted features emphasize different aspects, leading to variations in the feature representation across channels in $C_{2}$.  Notably, the proportion of tiny object features varies among different channels of $C_{2}$, which consequently affects their contribution to the final detection performance. The purpose of the channel branch is to identify channels in $C_{2}$ that contain more tiny object features and assign different weights to enhance their feature representation. 

\begin{figure}[t]
    \centering
    \includegraphics[width=0.95\columnwidth]{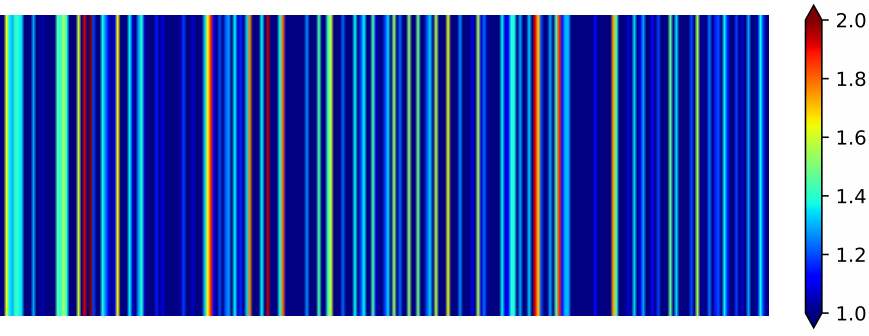} 
    \caption{Visualization of channel weight generated by HFP.}
    \label{fig13}
\end{figure}

However, since tiny object features occupy only a small proportion of the original feature map, computing channel weights directly on the entire feature map is susceptible to interference from low-frequency homogeneous background features. To address this, the High Frequency Generator (HFG) in Figure \ref{fig3} can be used to filter out most low-frequency components in $C_{2}$, resulting in the high-frequency response $F_{2}$. Since tiny object features belong to the high-frequency components, and the low-frequency components are removed in $F_{2}$, the proportion of tiny object features in $F_{2}$ is significantly increased compared to $C_{2}$. At this stage, we can more accurately identify the presence of tiny object features in each channel of $C_{2}$ based on $F_{2}$, thereby computing more precise channel weights. 

The channel weights generated by HFP are visualized in Figure \ref{fig13}, where the channel branch adaptively assigns different weights to each channel based on its high-frequency response. This ensures that the weights in the input feature map are no longer uniform but are dynamically adjusted according to the tiny object features present in each channel.

\begin{figure*}[t]
    \centering
    \includegraphics[width=0.95\textwidth]{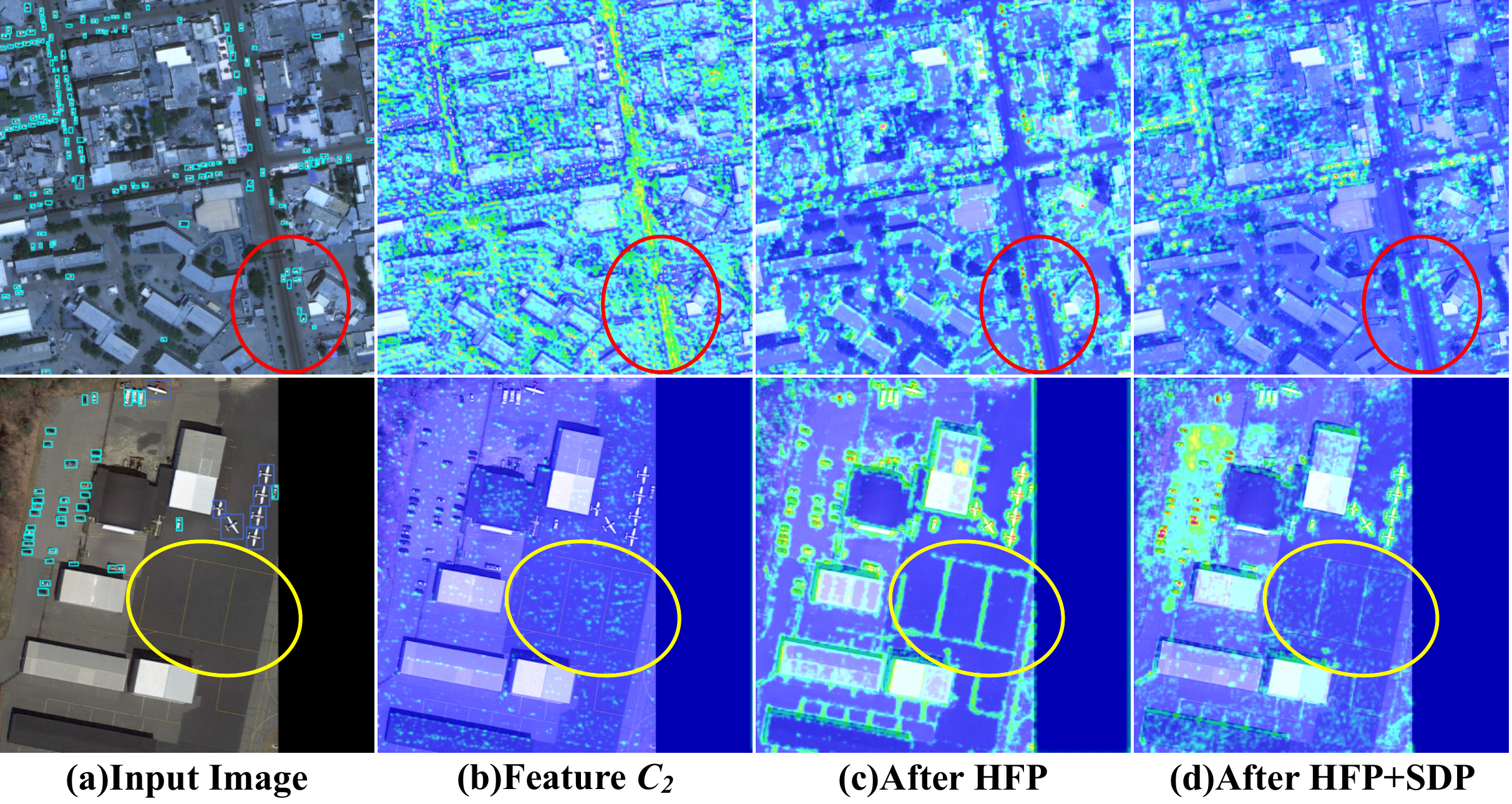} 
    \caption{Visualization of the feature at the $P_{2}$ level after HS-FPN.}
    \label{fig14}
\end{figure*}

\begin{figure}[t]
    \centering
    \includegraphics[width=0.99\columnwidth]{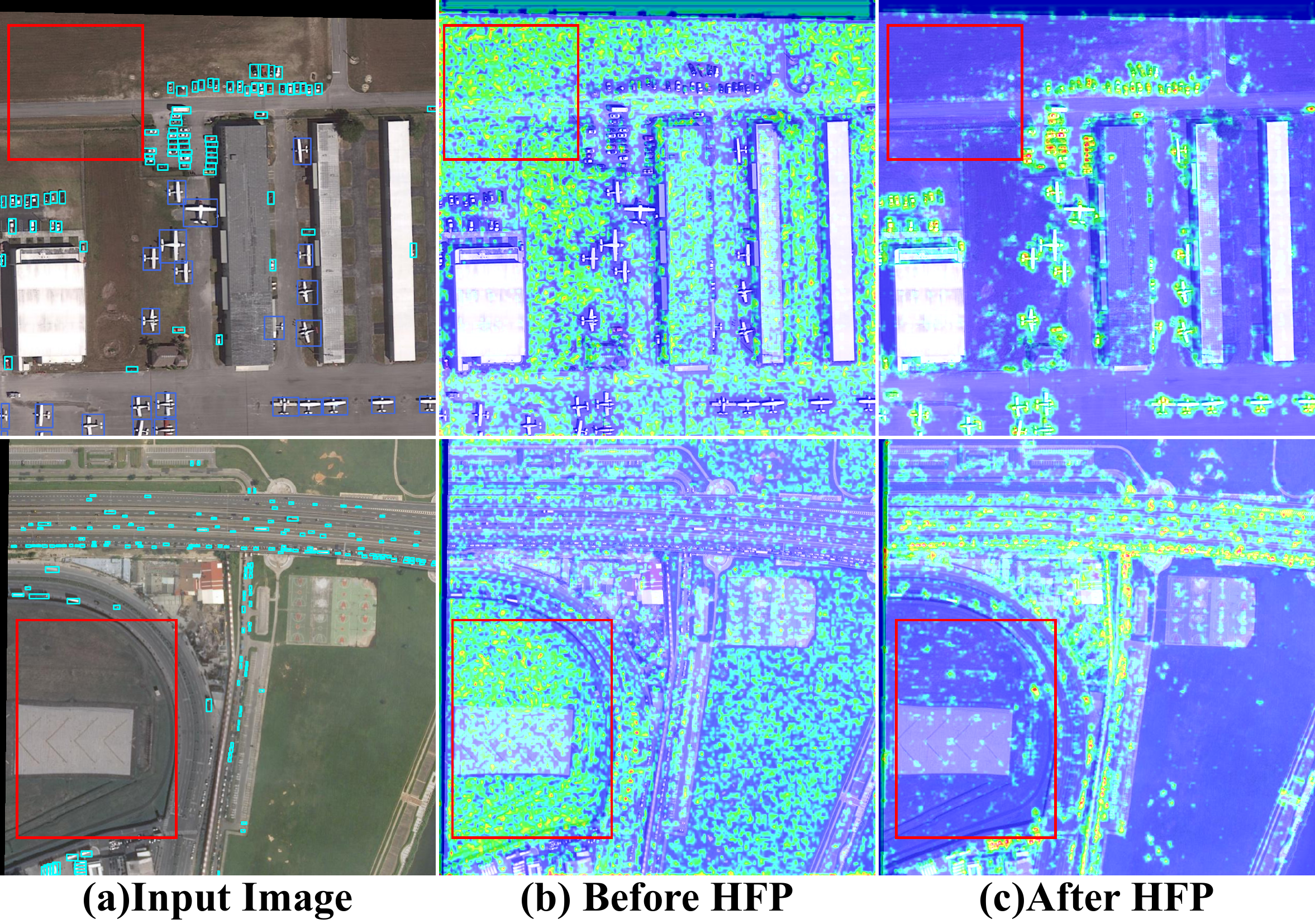} 
    \caption{Visualization of the feature at the $P_{2}$ level after HFP.}
    \label{fig15}
\end{figure}

\begin{figure}[t]
    \centering
    \includegraphics[width=0.99\columnwidth]{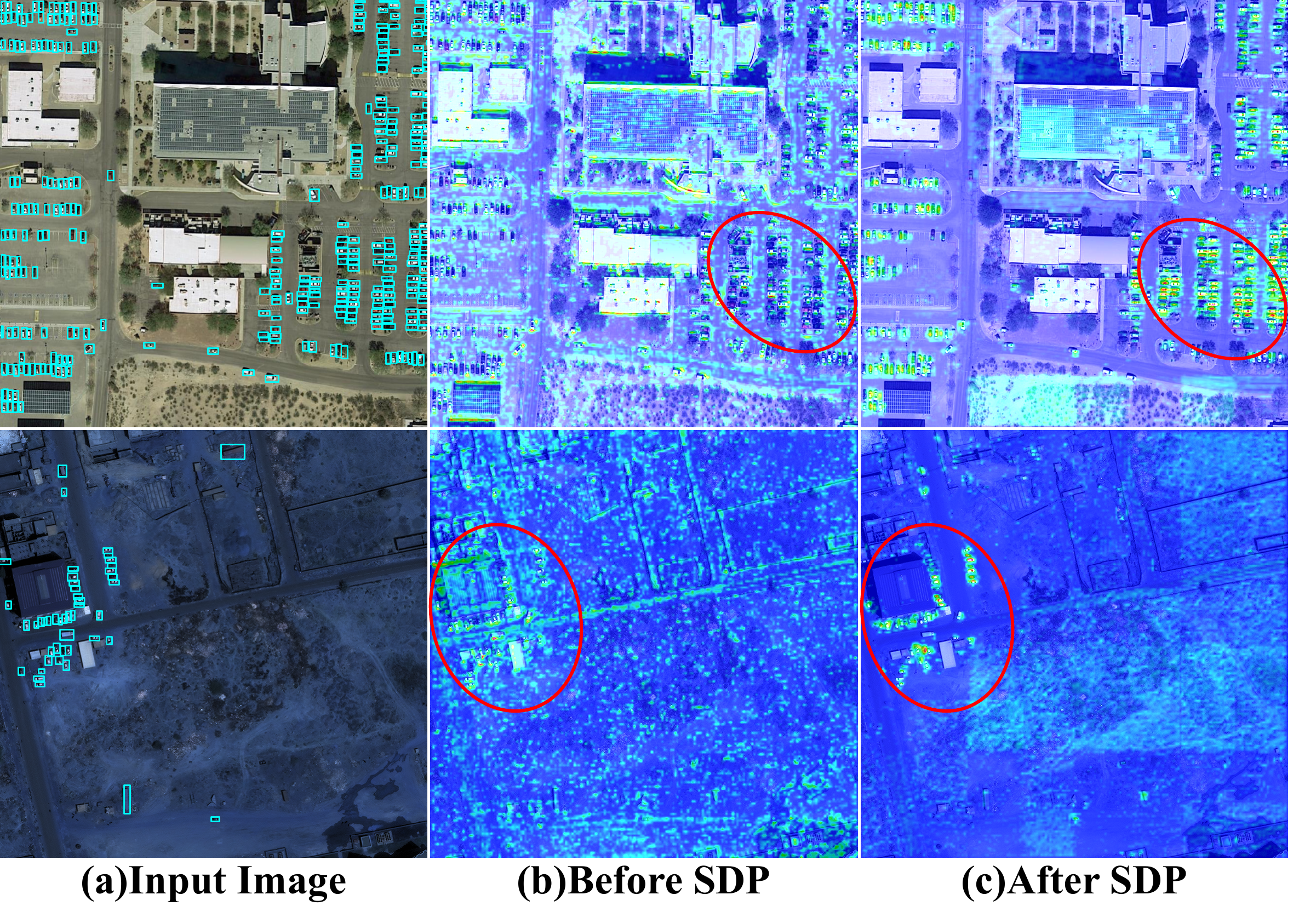} 
    \caption{Visualization of the feature at the $P_{2}$ level after SDP.}
    \label{fig16}
\end{figure}

Figures \ref{fig14}, \ref{fig15}, and \ref{fig16} respectively present the visualization results of the P2 layer features before and after processing by HS-FPN, HFP alone, and SDP alone. It can be observed that each component of HS-FPN effectively expands the feature dimensions of tiny objects and enhances their feature responses.

\subsection{Differences between SDP and ViT}
Given that the Spatial Dependency Perception module (SDP) in the main text uses an attention mechanism, there are some key differences from the attention mechanism used in ViT \cite{vit}. This section explains these differences in detail.

Figure \ref{fig16} illustrates the process of computing attention in both ViT and SDP. Assume the input feature has a shape of $R^{(c\times{H}\times{W})}$, with a patch size of $R^{(h\times{w})}$. Both ViT and SDP divide the input into $n$ feature blocks, where $n=\frac{H}{h}\times{\frac{W}{w}}$. Each feature block has a size of $R^{(c\times{h}\times{w})}$. The differences between the two methods are as follows:

As shown in Figure \ref{fig16} (a), ViT flattens each feature block into a one-dimensional vector of size $(1, hwc)$, effectively treating each feature block as a "point". ViT then computes the similarity between the $n$ "points" to produce a similarity matrix of size $R^{(n\times{n})}$. The computational complexity in ViT is $O({(n)}^2{hwc})$.

In contrast, as shown in Figure \ref{fig16} (b), SDP further divides each feature block into individual pixels, where each pixel is represented as a one-dimensional vector of size $(1, c)$. For each feature block, SDP calculates the similarity between pixels, resulting in $n$ similarity matrices, each of size $R^{((hw)\times{(hw)})}$. The computational complexity in SDP is $O(n(hw)^{2}c)$.

\begin{figure}[t]
    \centering
    \includegraphics[width=0.95\columnwidth]{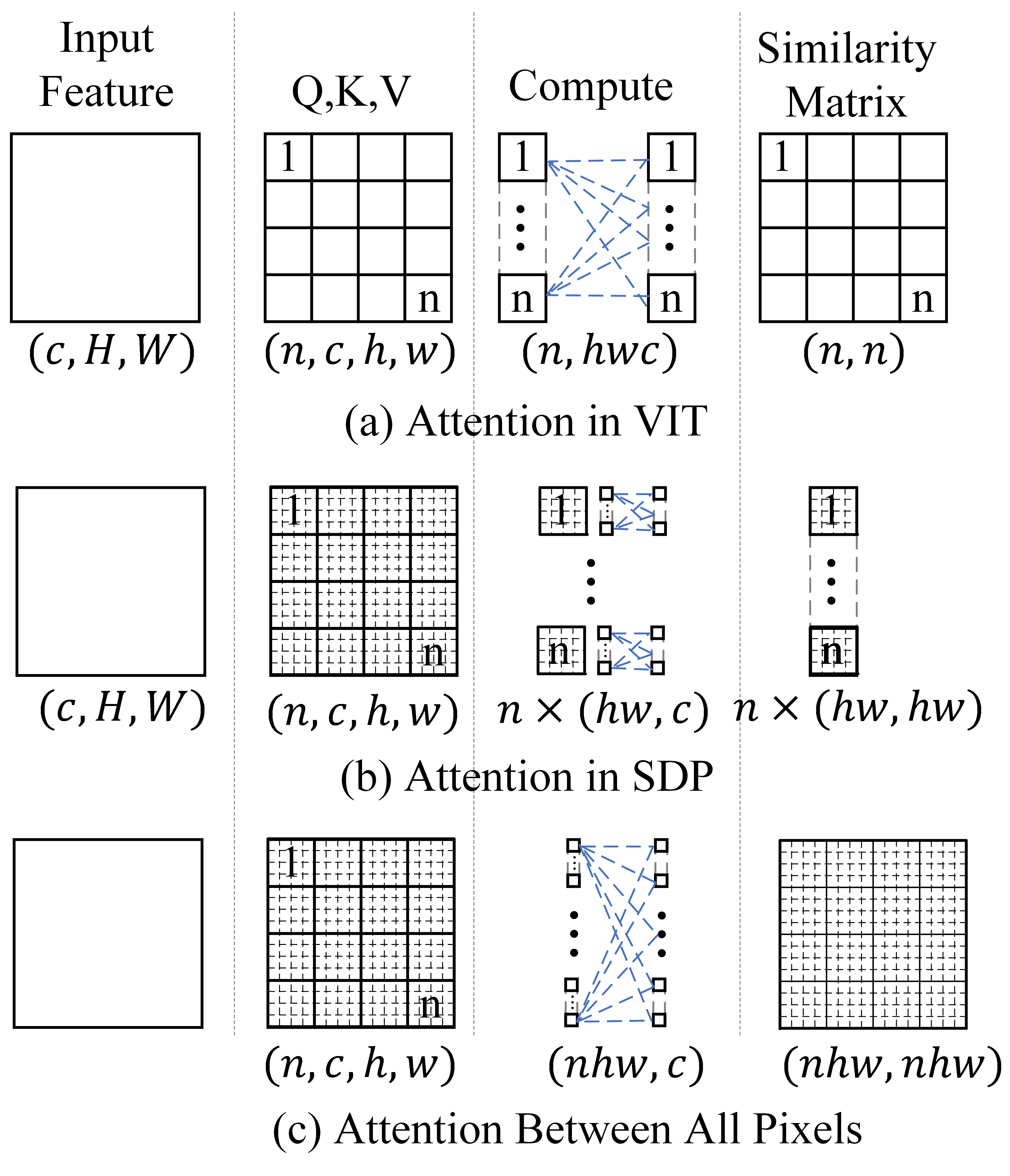} 
    \caption{Computing details between different attention methods. (a) Attention between feature blocks used in ViT; (b) Attention between pixels used in SDP; (c) Attention between all pixels.}
    \label{fig16}
\end{figure}

In summary, ViT implements region-level attention, whereas SDP focuses on pixel-level attention. As a result, SDP is capable of learning spatial dependencies between pixels and their local surroundings, thereby enriching the spatial context information of tiny objects. However, it is important to note that the computational cost of SDP is $(hw)/{n}$ times that of ViT, and this cost increases as the patch size $hw$ becomes larger.

Additionally, it is worth noting that the pixel-level attention in SDP is not applied globally, as doing so would significantly increase computational complexity. Figure \ref{fig16} (c) illustrates the process of calculating attention across global pixels, which results in a similarity matrix of size $R^{((nhw)\times{(nhw)})}$ and a computational complexity of $O((nhw)^{2}c)$. We believe that since the feature size of tiny objects is typically small, it is unnecessary to compute attention globally across all pixels.

\begin{table}[htbp]
  \centering
  \setlength{\tabcolsep}{1.5mm}{

    \begin{tabular}{lcc}
    \toprule
    Method & Complexity & {Multiplier} \\
    \midrule
    ViT   &$O({(n)}^2{hwc})$    & $1$  \\
    SDP   &$O(n(hw)^{2}c)$     &$\frac{(hw)}{n}$  \\
    Gloable  &$O((nhw)^{2}c)$     &${hw}$  \\
    \bottomrule
    \end{tabular}%
    }
  \label{tab7}%
  \caption{Computational complexity of different attention methods.}
\end{table}%

In conclusion, we consider the attention calculation method in SDP to be effective and the computational complexity to be within an acceptable range.

\subsection{Datasets and Metrics}
This section we will explain the motivation and building details about building $DOTA_{mini10}$ dataset.  

\textbf{Motivation.}
In order to train a robust model for detecting tiny objects, it is essential to have a large number of tiny instances. Currently, among the publicly available datasets for tiny object detection, we have primarily used the AI-TOD \cite{aitod} dataset for model training. This is because other datasets, such as AI-TOD-v2's \cite{aitod-v2} images are nearly identical to AI-TOD, and the SODA \cite{soda} dataset evaluates on entire large-scale images, resulting in very slow evaluating speeds. Although the DOTA \cite{dota} dataset is highly suitable for our research needs, it contains a significant number of large-scale instances and suffers from severe class imbalance (with the maximum class imbalance ratio being 77:1). Therefore, we have chosen to create a subset $DOTA_{mini10}$ from the DOTA dataset to make it more suitable for studying tiny object detection.

\begin{table}[htbp]
  \centering
  \setlength{\tabcolsep}{1.5mm}{
    \begin{tabular}{lr|r}
    \toprule
    \textbf{Category} & {\textbf{\#Instances}} & {\textbf{\#Instances}} \\
    \midrule
    plane & 28,373 & 7,168 \\
    small-vehicle & 84,248 & 16,026 \\
    large-vehicle & 57,569 & 13,278 \\
    ship  & 109,511 & 35,376 \\
    tennis-court & 8,512 & 1,654 \\
    baseball-diamond & 3,320 & 649 \\
    swimming-pool & 6,857 & 1,027 \\
    basketball-court & 3,130 & 407 \\
    storage-tank & 20,599 & 6,416 \\
    helicopter & 5,967 & 218 \\
    \midrule
    total & 328,086 & 82,219 \\
    \midrule
    image &10,653 &2,733 \\
    \bottomrule
    \end{tabular}%
    }
  \label{tab6}%
  \caption{Numbers of instances of each category of $DOTA_{mini10}$ train set (left) and validation set (right).}
\end{table}%

\begin{figure*}[t]
    \centering
    \includegraphics[width=0.95\textwidth]{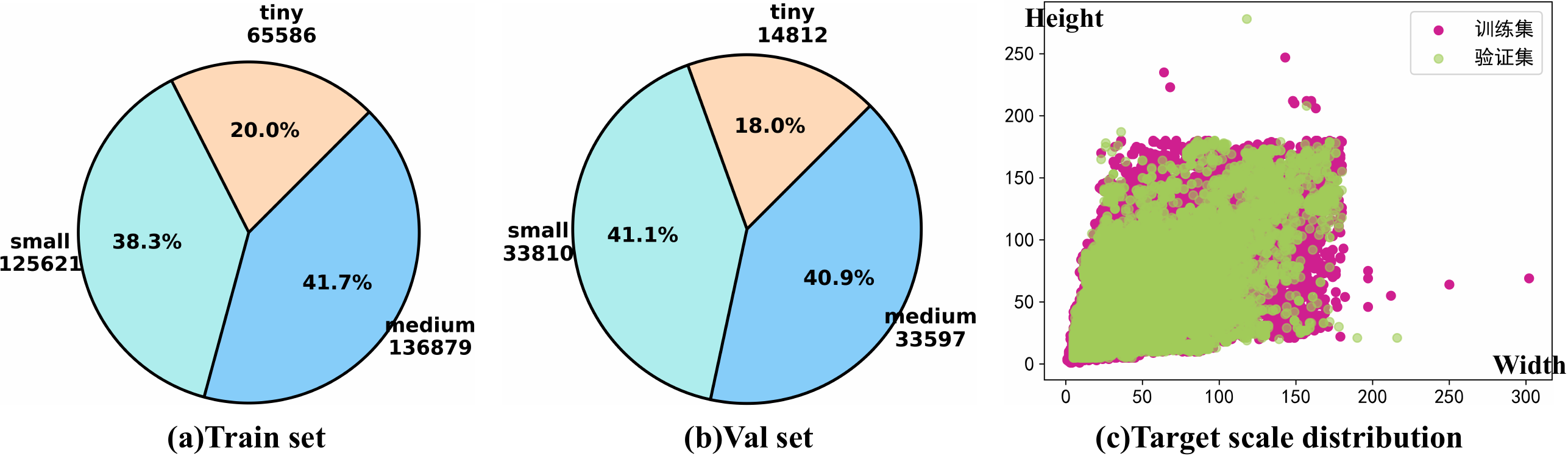} 
    \caption{Target size distribution map of $DOTA_{mini10}$.}
    \label{fig17}
\end{figure*}

\textbf{Building details.}
Firstly, We selected 10 balanced categories from the DOTA train set and validation set, including plane (PL), small-vehicle (SV), large-vehicle (LV), ship (SH), tennis-court (TC), baseball-diamond (BD), swimming-pool (SV), basketball-court (BC), storage-tank (ST), helicopter (HE). Second, we cleaned and organized the annotations of the original DOTA dataset. We then scaled the images in the dataset by a factor of 0.5 to 1.0, and applied parameters with a stride of 768, an overlap of 256 to split the images, with sub-images of size $1024\times1024$ pixels. Finally, we performed HSV color space transformation and probability flipping (p-flip) on the objects of the three categories: BC, BD, HE, in order to augment the number of instances. We named the dataset constructed above as $DOTA_{mini10}$, which consists of 10,653 images and 328,086 object instances in the train set, and 2,733 images and 82,219 object instances in the validation set. Notably, $17.3\%$ of instances are smaller than 16 pixels, and $38.5\%$ are smaller than 32 pixels in $DOTA_{mini10}$, respectively.

\textbf{Metrics.}
The evaluation metric used is the average precision ($AP$) metric provided in AI-TOD, following the COCO \cite{coco} style evaluation. In contrast to COCO, AI-TOD categorizes instances based on their scales into the following classes: objects within the range of 2 to 8 pixels are classified as very tiny, 8 to 16 pixels as tiny, 16 to 32 pixels as small, and above 32 pixels as normal-sized objects. These classes correspond to$AP_{vt}, AP_{t}, AP_{s}, AP_{m}$, respectively.

\subsection{Train Details}
\textbf{Model setting.} 
If not specifically mentioned, all models take the pre-trained ResNet50 \cite{resnet} as backbone network. Some models are also tested with ResNet101 and MobileNetv2 \cite{mobile} for comparison. Additionally, the number of proposals in the region proposal network (RPN) is set to 3000, and the feature pyramid network (FPN) has 256 channels. During the inference stage, a confidence score threshold of 0.05 is applied to filter out background bounding boxes, and non-maximum suppression (NMS) with an IoU threshold of 0.5 is used to generate the top 3000 confident bounding boxes. The most important thing is that due to the presence of a large number of tiny objects (less than 16 pixels) in AI-TOD and $DOTA_{mini10}$, the conventional anchor sizes are not suitable. Therefore, for anchor-based models, we set the anchor base size to 2 to ensure that an adequate number of positive samples are generated for training tiny objects.

\subsection{Visualization Results.}
\begin{figure*}
    \centering
    \includegraphics[width=0.8\textwidth]{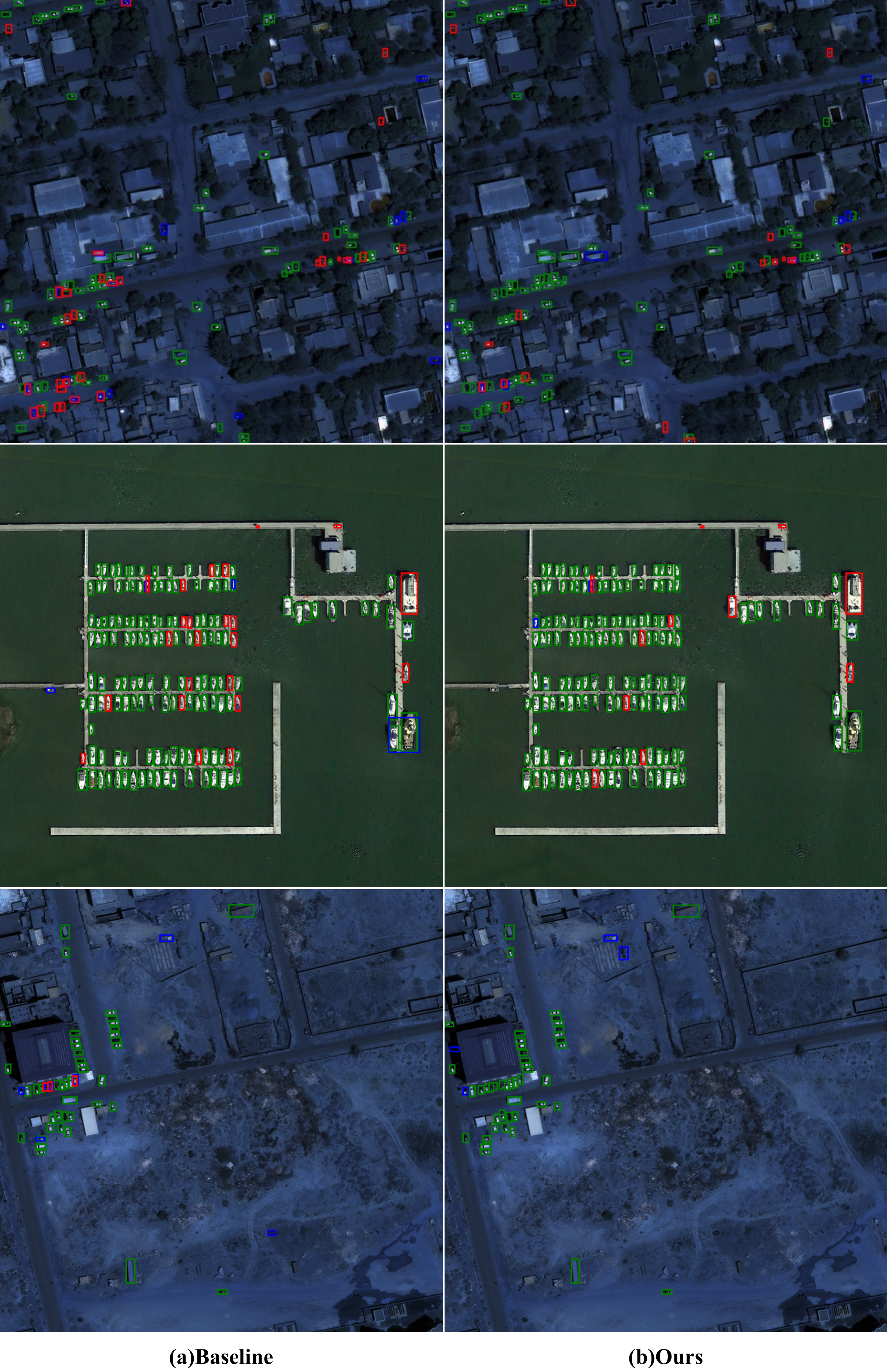}
    \caption{Visualization results on AI-TOD. The first and third rows are the results of Cascade R-CNN with FPN, the second and fourth rows are the results of Cascade R-CNN with HS-FPN. The green, blue and red boxes in the figure donate true positive (TP), false positive (FP) and false negative (FN) predictions. Zooming in for more details.}
    \label{fig18}
\end{figure*}

\begin{figure*}
    \centering
    \includegraphics[width=0.8\textwidth]{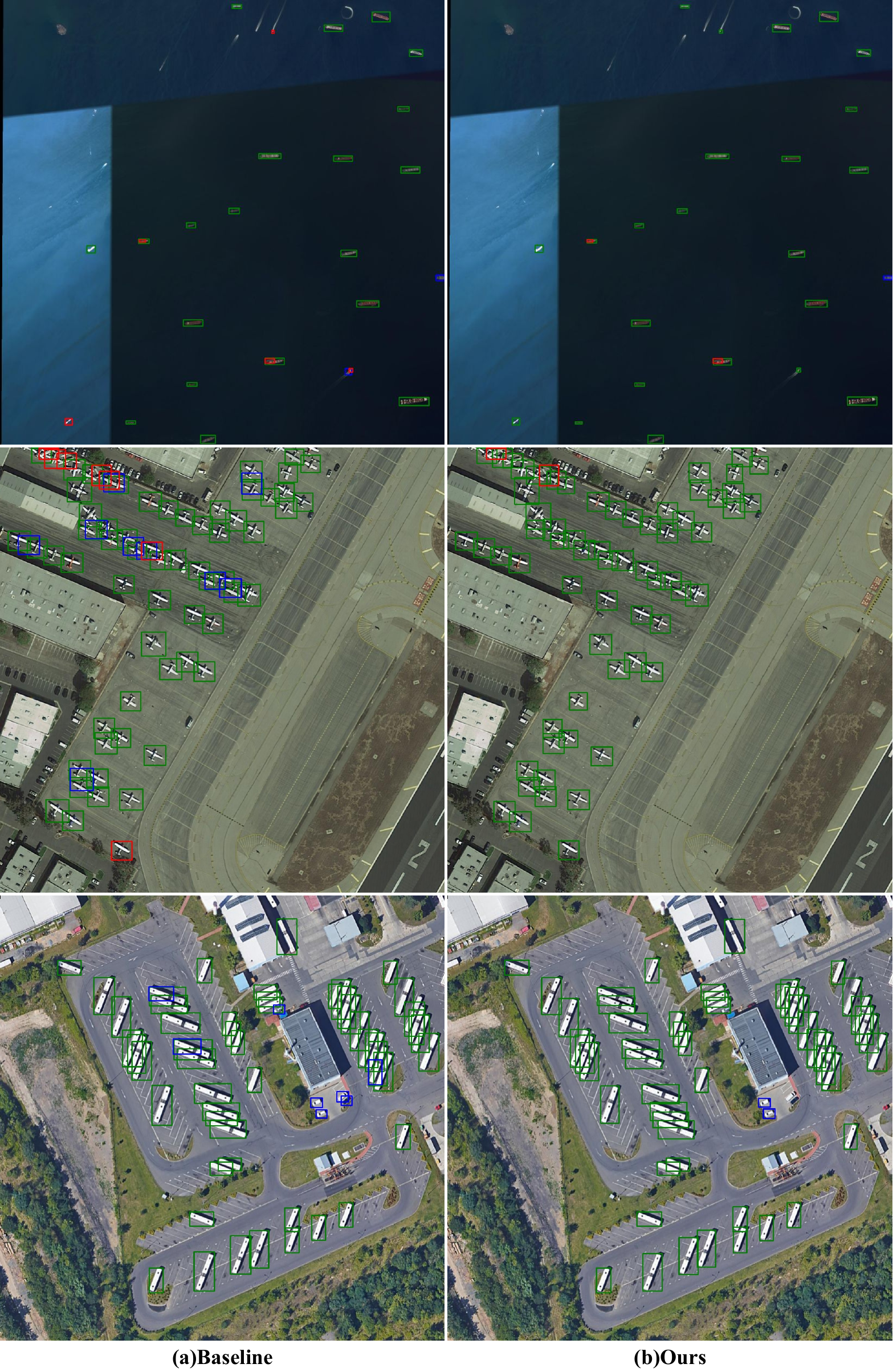}
    \caption{Visualization results on $DOTA_{mini10}$. The first and third rows are the results of Cascade R-CNN with FPN, the second and fourth rows are the results of Cascade R-CNN with HS-FPN. The green, blue and red boxes in the figure donate true positive (TP), false positive (FP) and false negative (FN) predictions. Zooming in for more details.}
    \label{fig19}
\end{figure*}

Figure \ref{fig18}, \ref{fig19} shows the visualization results on AI-TOD and $DOTA_{mini10}$. The first and third rows are the results of Cascade R-CNN with FPN, the second and fourth rows are the results of Cascade R-CNN with HS-FPN. Backbones are both ResNet50. It can be observed that replacing FPN with HS-FPN significantly reduces both FN and FP predictions for the targets, indicating that HS-FPN improves the detection performance of tiny objects.

\end{document}